\documentclass[11pt]{article}

\usepackage[T1]{fontenc}
\usepackage[utf8]{inputenc}

\DeclareUnicodeCharacter{2265}{\ensuremath{\geq}}
\DeclareUnicodeCharacter{2264}{\ensuremath{\leq}}
\DeclareUnicodeCharacter{2208}{\ensuremath{\in}}
\DeclareUnicodeCharacter{2260}{\ensuremath{\neq}}
\DeclareUnicodeCharacter{2248}{\ensuremath{\approx}}

\usepackage{amsmath,amsthm}
\usepackage{mathtools}

\usepackage{XCharter}
\usepackage[xcharter]{newtxmath}

\usepackage{bm}

\usepackage[round]{natbib}
\usepackage{xcolor}
\usepackage{graphicx}
\graphicspath{{./}{../}}
\usepackage{booktabs}
\usepackage{adjustbox}
\usepackage{subcaption}
\usepackage{multirow}

\providecommand{\TextWidth}{6.0in}
\providecommand{\TextHeight}{9.0in}
\usepackage[letterpaper]{geometry}
\usepackage{microtype}
\theoremstyle{plain}

\theoremstyle{definition}

\usepackage{fancyhdr}
\usepackage{titlesec}
\titleformat{\section}
  {\normalfont\Large\bfseries\scshape}{\thesection}{0.7em}{}
\titlespacing*{\section}{0pt}{2.4ex plus 1ex minus .2ex}{1.2ex plus .2ex}
\titleformat{\subsection}
  {\normalfont\large\bfseries}{\thesubsection}{0.6em}{}
\titlespacing*{\subsection}{0pt}{1.9ex plus .8ex minus .2ex}{0.8ex plus .2ex}

\definecolor{linknavy}{HTML}{1A4D8F}
\usepackage{hyperref}
\hypersetup{
  colorlinks=true,
  linkcolor=linknavy,
  citecolor=linknavy,
  urlcolor=linknavy,
  pdftitle={Measuring Dead Directions: Decomposing and Classifying Singular Structure off Canonical Alignment},
}
\usepackage{orcidlink}
\usepackage{etoc}       

\usepackage{titling}
\pretitle{\begin{center}\LARGE\bfseries\scshape}
\posttitle{\par\end{center}\vskip 0.45em {\centering\rule{\linewidth}{0.8pt}\par}\vskip 0.7em}
 
\newcommand{\fisher}{F}

\newcommand{\cfc}{c_{\mathrm{fc}}}
\newcommand{\cproj}{c_{\mathrm{proj}}}

\title{Measuring Dead Directions:\\
Decomposing and Classifying Singular Structure\\
off Canonical Alignment}
\author{
  Tejas Pradeep Shirodkar\thanks{Correspondence: \texttt{tejas.shirodkar@research.iiit.ac.in} \quad
    \orcidlink{0009-0001-3034-0087}\,\href{https://orcid.org/0009-0001-3034-0087}{0009-0001-3034-0087}} \\
  IIIT, Hyderabad
}
\date{}

\begin{document}
\maketitle

\begin{abstract}
We give a descent-free, alignment-free measurement of singular structure on trained
networks. At a single frozen checkpoint the read recovers the order $k$ of each dead
direction from the directional-Fisher rate, the master invariant from which the
per-direction learning coefficient $1/(2k)$ follows exactly, in whatever basis the
optimizer left. The same read classifies each direction, separating a genuine
singularity, whose order the architecture fixes, from a flat gauge symmetry; the
directional-Fisher magnitude settles the cases the order cannot. A pluggable detector
supplies the directions for transformer, convolutional, and normalisation layers. The
read recovers the architecture-predicted order across constructed cells and trained
networks, including a fine-tuned vision transformer whose dead structure is the
LayerNorm-kernel gauge and a from-scratch one whose compressed MLP forms a node-death
at its activation order. Where the singular structure enumerates, the per-direction
orders assemble, through the typed intersection of the loci, into the global coefficient
$(\lambda, m)$ matching the closed form. The method removes the canonical-alignment and
descent preconditions of the underlying rate result, turning order-recovery into a
deterministic, architecture-general reading. We then map its reach into the Watanabe
triple: the order determines the universal singular fluctuation $\nu(k)$, though a
trained network's realized $\nu$ falls below it as the live structure absorbs the dead
direction's data fluctuation, and the multiplicity recovers from the dominant structure
under a single-locus assumption.
\end{abstract}

\etocsettocdepth.toc{none}

\section{Introduction}
\label{sec:intro}

A dead direction is the object two traditions see at the same point. From Amari's
information geometry \citep{Amari16} it is a direction in which the Fisher metric loses
non-degeneracy. From Watanabe's singular learning theory \citep{Watanabe09} it is tangent to the
analytic singular set, where the Kullback--Leibler divergence vanishes to an integer order that
resolution of singularities recovers. The two readings name the same vector, and that order $k$
is the invariant that bridges them. The trajectory-rate result of \citet{TheoryRefNamed} reads
$k$ in original parameter coordinates, without resolution: move the parameters along a dead
direction $u$, $\theta(t)=\theta_0+tu$, and the directional Fisher decays as
$u^\top \fisher(\theta(t))\,u = \Theta(t^{2(k-1)})$, so the log-log slope returns $k$
and the per-direction learning coefficient $\lambda = 1/(2k)$.

\begin{figure}[t]
\centering
\includegraphics[width=\textwidth]{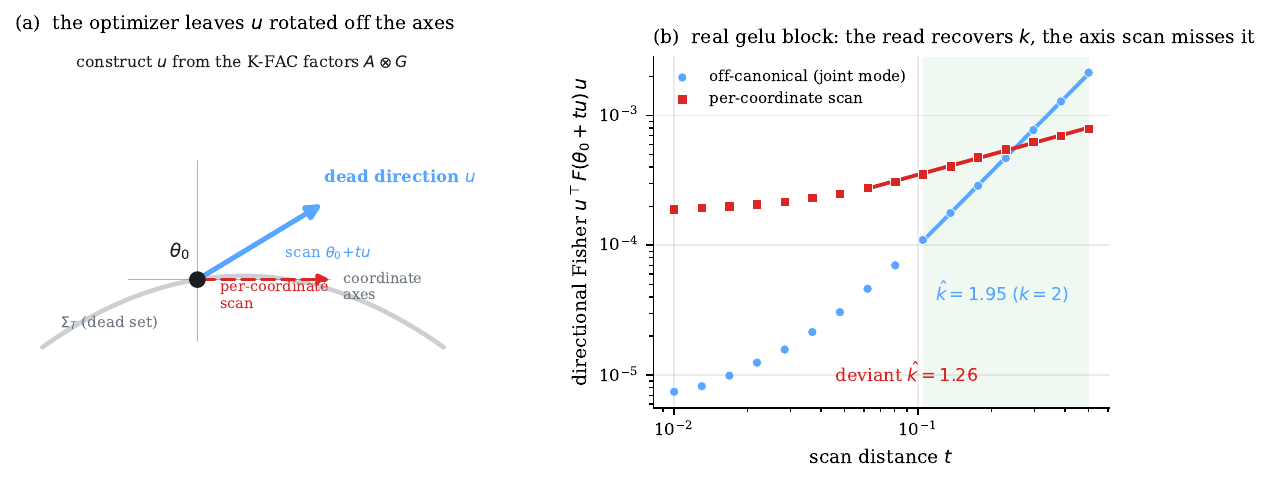}
\caption{Reading the order off canonical alignment. (a) A trained network leaves a dead
direction $u$ rotated off the coordinate axes; we construct $u$ as the joint mode of the K-FAC
factors $A\otimes G$ and scan the directional Fisher out from the frozen checkpoint $\theta_0$,
with no descent and no alignment. (b) On a real gelu transformer block whose dead direction is
rotated off the axes, the off-canonical joint-mode read recovers the activation order
($\hat k=1.95$, $k=2$), while a per-coordinate scan along an axis follows the wrong direction and
reads a deviant order ($\hat k=1.26$). The slope in the purity-matched window (shaded) returns
$k$.}
\label{fig:hero}
\end{figure}

Singular learning theory characterises a trained network by this learning
coefficient together with the multiplicity $m$ and the singular fluctuation $\nu$,
the Watanabe triple $(\lambda, m, \nu)$ that controls the Bayesian free energy and the widely
applicable information criterion (WAIC) \citep{Watanabe18}. Reading the triple on a real network is costly and
preconditioned. The standard estimator of the learning coefficient samples the
posterior with stochastic-gradient Langevin dynamics (SGLD) \citep{LauFurmanWangMurfetWei25}; it returns a single calibrated scalar, requires
per-model tuning of the sampler, does not localise to a network coordinate, and does
not isolate $m$ or $\nu$. The rate read above is cheaper, but the clean per-layer
version of \citet{TheoryRefNamed} assumes canonical alignment, the dead direction
being the same coordinate at every layer, and a descending, theorem-compatible
optimizer. Trained networks meet these conditions only in part.

We give a measurement methodology that applies wherever a dead direction has formed, the
regime where a network carries singular structure to read. Given one, the pipeline is
\emph{detect then read}: a detector locates
the direction, and one descent-free scan at a single frozen checkpoint reads its
order from the directional Fisher rate, with a purity-matched window isolating the
$t^{2(k-1)}$ regime. The read returns the order $k$ per direction, hence
$\lambda_{\mathrm{dir}} = 1/(2k)$, with the dead-subspace dimension alongside, so the single coefficient that
posterior sampling reports resolves into the per-direction structure it sums over.
The scan also reads which kind of direction it found. A finite order marks a genuine
degeneracy: a \emph{node-death}, a hidden unit whose incoming and outgoing weights have both
collapsed, whose order is the unit activation's local analytic order, or a depth-induced
singularity of a deep linear map, whose order is the depth. A
directional Fisher that stays at the floor marks a gauge direction, a symmetry of the
architecture that adds to the multiplicity without carrying a finite order.
Separating the two takes the magnitude of the Fisher: a curved gauge orbit read along its
tangent imitates a finite order on the slope alone. We exercise the read across a taxonomy of
dead-structure types, from constructed node-deaths and deep-linear depth singularities to a
real fine-tuned vision transformer's LayerNorm-kernel gauge and a from-scratch one's rotated
node-death (Section~\ref{sec:taxonomy}).

The detector is the part of the pipeline that changes with the architecture, since
the structure that exposes a dead direction changes with it. A generic layer surfaces
the direction as a near-kernel of a K-FAC (Kronecker-factored approximate curvature)
factor \citep{MartensGrosse15}; the detector forms the activation factor and the gradient factor and reads the
direction off whichever separates it more cleanly, the activation--gradient duality of
\citet{TheoryRefNamed}. A convolutional layer replaces the activation factor with a
spatial-patch covariance, and the read runs on that. A LayerNorm transformer needs no
scan, since the kernel of its normalisation scale gives the direction in closed form.
The read downstream is identical in each case.

The read and the posterior sampler are complementary. The read is deterministic and needs no
canonical alignment or descent, and the sampler supplies the single calibrated coefficient the
read decomposes into per-direction structure. Beyond the order and its exact coefficient
$1/(2k)$, the read reaches the universal fluctuation $\nu(k)$ through the order and the
multiplicity through the dominant structure, though a trained network realizes both only in part.

\paragraph{Contributions.}
\begin{enumerate}
\item A descent-free, alignment-free read of the per-direction order $k$, hence
$\lambda_{\mathrm{dir}} = 1/(2k)$, and of the dead-subspace dimension, at a single frozen checkpoint
(Figure~\ref{fig:hero}), through
duality-based identification, by constructing the dead mode from the factors rather
than searching the Fisher spectrum for it, and a purity-matched rate window
(Section~\ref{sec:method}).
\item A detect-then-read pipeline whose detector adapts to the architecture, a K-FAC
dual-factor scan, a convolutional channel-death factor, or the algebraic
LayerNorm-kernel direction, while the read stays fixed (Section~\ref{sec:method}).
\item A taxonomy that classifies each dead direction as a genuine singularity, whose
finite order the architecture fixes (a node-death at the activation order, a depth
singularity at the network depth, a unit-overlap merging), or a flat gauge that adds to
the multiplicity without an order (the LayerNorm-kernel, attention-rotation, and
cross-entropy-shift gauges), with the magnitude criterion that tells a curved gauge orbit
from a finite order. We populate it across constructed cells, a from-scratch vision
transformer whose compressed MLP forms a genuine node-death, and a real fine-tuned one
whose dead structure reads as the LayerNorm-kernel and attention-rotation gauges
(Section~\ref{sec:taxonomy}, Table~\ref{tab:taxonomy}).
\item The optimizer-dependent geometry the read reports: a standard optimizer can leave a
deep network's dead structure too diffuse to carry an order, or a shallower network's rotated
off the coordinate axes, where a per-coordinate scan misses it, while an orthogonalising optimizer forms a
clean dead structure the off-canonical read recovers (Section~\ref{sec:optimizer}).
\item The global coefficient where the singular structure enumerates: the per-direction
orders assemble, through the typed intersection of the loci (transversal, separable,
tangency, or determinantal), into the global $(\lambda, m)$ matching Aoyagi's closed form
on the analytic cells, set beside the posterior sampler where enumeration is open
(Sections~\ref{sec:comparison}, Appendix~\ref{app:global}).
\item A map of the read's reach into the rest of the Watanabe triple: the order
determines the universal singular fluctuation $\nu(k)$, which we confirm by sampling on an
isolated order-$k$ direction; a trained network's realized $\nu$ falls below $\nu(k)$ as
the live structure absorbs the dead direction's data fluctuation, a suppression we isolate
and measure; and the multiplicity recovers from the dominant structure under a
single-dominant-locus assumption (Section~\ref{sec:trajectory}).
\end{enumerate}
 \section{Background}
\label{sec:background}

\paragraph{The rate primitive.} Along a dead direction the directional Fisher decays as
$t^{2(k-1)}$ for the direction's KL order $k$, a measure of how flat the loss is along the
direction: the larger $k$, the more derivatives vanish before the loss responds. The order is
the invariant both traditions read in original coordinates \citep{TheoryRefNamed,Watanabe09}. That order fixes the direction's
local threshold $\lambda_{\mathrm{dir}}=1/(2k)$. The global threshold collects the
per-direction contributions, summing $\sum_i 1/(2k_i)$ over independent dead directions but
taking $\min_i 1/(2k_i)$ where directions meet in a normal crossing (their singular loci
intersect transversally), so a global value needs both the per-direction orders and the
crossing structure that combines them.

\paragraph{The Watanabe triple.} A regular model, one with an identifiable parameter and a
non-degenerate Fisher metric, has its Bayesian free energy and generalization error set by half
the parameter count, $d/2$. Neural networks are singular: distinct parameters realize the same
function, and the Fisher degenerates on the set $\Sigma_T$ of optimal parameters, so $d/2$ no
longer applies \citep{Watanabe09}. Singular learning theory replaces it with three invariants of
that singular structure. The learning coefficient, or real log canonical threshold, $\lambda$ is
the effective complexity: it is the coefficient of the leading correction to the free energy
$F_n = nL_0 + \lambda\log n - (m-1)\log\log n + O(1)$, equals $d/2$ for a regular model, falls
below it as the structure grows more degenerate, and governs the Bayes generalization error. The
multiplicity $m$ counts the components of $\Sigma_T$ that achieve this minimal $\lambda$, and
sets the $\log\log n$ term. The singular fluctuation $\nu$ governs the gap between generalization
and training loss, through the widely applicable information criterion
$\mathrm{WAIC}=T_n+2\nu/n$ \citep{Watanabe18}, and also reduces to $d/2$ in the regular case. For
the analytic singular models, reduced-rank regression and deep linear networks, the triple is
known in closed form \citep{AoyagiWatanabe05,Aoyagi24}; these are the ground truth the paper
calibrates against.

\paragraph{The two preconditions.} The clean per-layer rate read of \citet{TheoryRefNamed}
holds under two conditions that a trained network meets only in part: \emph{canonical
alignment}, so that the dead direction occupies one coordinate at every layer and a
per-coordinate scan follows it, and \emph{descent} under a theorem-compatible optimizer, so
that the rate is read along the approach to the singularity. We keep neither, asking only
that a dead direction has formed and reading it at one frozen checkpoint in whatever basis
the optimizer leaves. In place of the descent, the read scans a synthetic displacement
$\theta_0+tu$ out from the checkpoint along the nominated direction $u$, with the scale $t$
set by the read, and takes the order from the growth of the directional Fisher along that
scan. No training trajectory enters; the checkpoint need only sit at the singularity in $u$,
the one precondition that remains.

\paragraph{K-FAC factors and the A--G duality.} The parameter Fisher of a layer $y=Wx$
factorises \citep{MartensGrosse15} as $\fisher_W\approx A\otimes G$, with input covariance
$A=\mathbb{E}[xx^\top]$ and output-gradient covariance $G=\mathbb{E}[gg^\top]$ for
$g=\partial L/\partial y$, and its smallest-Fisher direction is the rank-one lift
$g_{\min}a_{\min}^\top$. The two factors are dual, with
$\lambda_{\min}(A_\ell)\,\lambda_{\min}(G_\ell)=\Theta(t^{2(L-1)})$ the same at every layer
$\ell$ of the depth-$L$ network \citep[Thm.~3]{TheoryRefNamed}, so a detector can read the
dead direction off whichever factor separates it at a given layer. We accumulate both factors as the \emph{true}
Fisher (label-resampled and Monte-Carlo estimated, \texttt{true-MC} in the tables), resampling the labels from the model's own predictive distribution. The
\emph{empirical} Fisher, built from the data labels, carries the model's fit error: there
$g$ is the residual at the observed label, which vanishes only at a perfect fit and
contributes no curvature. Resampling removes that term, so $\lambda_{\min}$ and the
effective rank \citep{RoyVetterli07} stay geometric quantities. The loss-landscape
degeneracy line of \citet{BushnaqMendel24} reads the same activation and gradient structure
at leading order and bounds the higher-order content by the Hessian rank, while the read
here scans the nominated direction for the order $k$ that the rank leaves unmeasured.

\paragraph{Node-death.} A hidden unit whose incoming weights $\cfc[j,:]$ and outgoing
weights $\cproj[:,j]$ have both collapsed contributes nothing to the output, the
fully-dead-unit configuration that singular learning theory studies as the canonical
two-layer singularity \citep{Carroll21,FarrugiaRoberts22,FarrugiaRoberts23}; we call it a
\emph{node-death}. Scaling the two halves together makes the unit's contribution bilinear,
so the output grows as $t^{k}$ and the directional Fisher as $t^{2(k-1)}$ with $k$ the
activation's local analytic order, $3$ for squared-ReLU and $2$ for gelu or ReLU. Scaling
one half alone leaves the function unchanged, a \emph{gauge} direction of the
hidden-reparametrisation symmetry $W_\ell\mapsto MW_\ell,\ W_{\ell+1}\mapsto
W_{\ell+1}M^{-1}$, joined by the LayerNorm centering symmetry. These gauge directions
occupy the Fisher's near-zero floor \citep[Thm.~3]{TheoryRefNamed} and carry no order. At a
frozen checkpoint we count the dead subspace, the genuine node-deaths at the dominant order,
excluding the gauge directions and the below-floor numerical near-kernels. Under a normal
crossing of equal-order directions this count is the analytic multiplicity $m$ (the constructed
overlap cell returns $m=n-1$); on a trained network it is the dead-subspace dimension, which
equals $m$ only under that crossing assumption and otherwise tracks the compressing subspace as
training proceeds. We report the dead-subspace dimension on the trained networks and reserve
$m$ for the analytic value we verify on the constructed cells.

\paragraph{Related work.} The standard estimator of the learning coefficient is the SGLD local
learning coefficient \citep{LauFurmanWangMurfetWei25}, extended to a weight subset by the refined
LLC \citep{WangHoogland24} and across training checkpoints by the stagewise-development reading
\citep{HooglandWangFarrugiaRoberts24}. The geometry read is the deterministic, sampling-free
counterpart of that family: it returns the per-direction order at a frozen checkpoint and is set
beside the sampler in Section~\ref{sec:comparison}, and its developmental tracking
(Appendix~\ref{app:developmental}) reads without sampling the same structure the stagewise LLC
follows. The loss-landscape degeneracy programme of \citet{BushnaqMendel24} diagonalises the
same K-FAC factors $A$ and $G$ to expose a sparsely interacting feature basis; the read here
scans those factors instead for the order the Hessian rank leaves unmeasured. The wider
singular-learning landscape in deep learning is surveyed by the theory paper \citep{TheoryRefNamed}.

This paper is the off-canonical measurement layer of one programme. It builds on the rate
primitive, the activation--gradient duality, the algebraic LayerNorm-kernel direction, and the
gauge quotient of the theory paper \citep{TheoryRefNamed}, the cheap spectral observables of
\citet{shirodkar2026deaddirectionsignaturescheapspectral}, the algebraic LayerNorm detector of
\citet{shirodkar2026algebraicdeaddirectionslayernorm}, and the gauge-equivariant optimizer of
\citet{DDCRef}. On top of these it removes the canonical-alignment and descent preconditions of
the rate read, classifies each dead direction as a genuine degeneracy or a flat gauge, and
decomposes the learning coefficient into the per-direction order and the dead-subspace dimension
at a frozen checkpoint. Its experiments re-analyse the modular-addition cohort of
\citet{DDCRef} and reuse the dead-unit census of
\citet{shirodkar2026deaddirectionsignaturescheapspectral}, reading at a frozen checkpoint what
those papers read in training and in the spectrum.

\paragraph{Notation.} Table~\ref{tab:notation} collects the symbols the read uses.

\begin{table}[t]
\centering\small
\caption{Notation. One symbol, $\gamma$, carries two meanings the context separates: the
LayerNorm per-channel gain in the gauge read, and the localization strength of the posterior
sampler in the global view.}
\label{tab:notation}
\begin{tabular}{ll}
\toprule
symbol & meaning \\
\midrule
$k$ & KL order of a dead direction (flatness of the loss along it) \\
$\lambda_{\mathrm{dir}}=1/(2k)$ & per-direction local threshold \\
$\lambda$ & global learning coefficient (RLCT) \\
$m$ & analytic multiplicity (directions at the dominant order); verified on constructed cells \\
$N_{\mathrm{dead}}$ & dead-subspace dimension, the Fisher-bottom count on a trained net \\
$\nu$ & singular fluctuation, $\mathrm{WAIC}=T_n+2\nu/n$ \\
$\theta_0$ & frozen checkpoint the read scans from \\
$u$ & nominated dead direction \\
$t$ & scan displacement scale, $\theta_0+tu$ \\
$\alpha$ & log-log slope of the directional Fisher, $\alpha=2(k-1)$ \\
$r^2$ & fit quality of the rate window (admissible at $r^2>0.95$) \\
$\fisher$, $A$, $G$ & Fisher, and its K-FAC input / output-gradient factors \\
$L$, $d$, $n$ & network depth, layer width, sample count \\
$\sigma_{\min}$ & smallest singular value of the dead direction \\
$\gamma$ & LayerNorm gain (gauge read) / sampler localization (global view) \\
\bottomrule
\end{tabular}
\end{table}
 \section{Method}
\label{sec:method}

\begin{figure}[t]
\centering
\includegraphics[width=\textwidth]{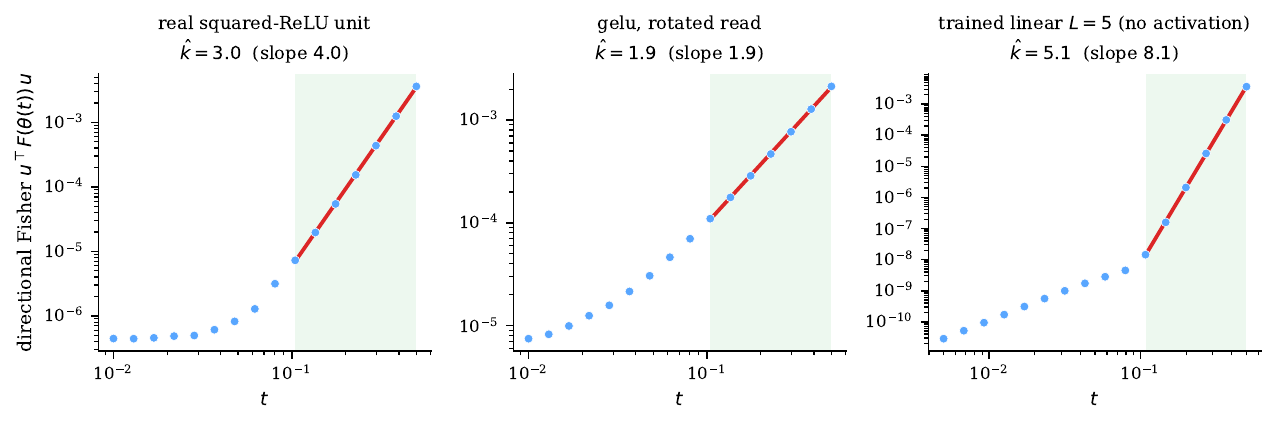}
\caption{The read in action, on a log-log directional Fisher against $t$ with the
purity-matched window shaded. (a) a real squared-ReLU dead unit, axis-aligned, returning
$k{=}3$; (b) a real gelu network's rotated dead direction, recovered off canonical
alignment at $k{=}2$; (c) a trained deep linear network with no activation, returning the
depth order $k{=}L{=}5$. In each the Fisher flattens onto a contamination floor as
$t\to0$ and follows the $t^{2(k-1)}$ power law in the window.}
\label{fig:rate_read}
\end{figure}

We measure the order of a dead direction at a frozen checkpoint, in whatever basis the optimizer
left, by a procedure in two stages. The detector reads the dead coordinate at each layer and
assembles the per-layer coordinates into one network-wide direction $u$; the read then scans the
directional Fisher along $u$ out from the checkpoint and takes the order from its growth
(Figure~\ref{fig:rate_read}). The two stages separate a coordinate-free quantity from an
architecture-specific one. The order is an analytic invariant of the singularity, the same number
in any basis, so the scan that recovers it is one fixed operation. Finding the direction it lives
along depends on where each layer type seats its dead coordinate, so only the detector varies with
the architecture. Three mechanisms make the measurement hold off
canonical alignment, taken in turn below: the duality that carries the dead coordinate across
layers, the construction that seats the scan on the order-carrying mode above the gauge floor, and
the purity-matched window that reads the exponent from a direction recovered only approximately.
The full pipeline and setup are Appendices~\ref{app:pipeline} and~\ref{app:setup}.

\paragraph{The detector across architectures.} The detector generalises by one rule: read the
dead coordinate as the near-kernel of the layer's natural second-moment object. A near-kernel is
the subspace of that object's smallest eigenvalues, the directions along which the metric it
builds has gone near-singular because the network stopped using them. For a generic
layer that object is the K-FAC factor pair $A\otimes G$ of Section~\ref{sec:background}; a
convolution replaces the input factor with its spatial-patch covariance \citep{GrosseMartens16};
a LayerNorm transformer needs no scan at all, its dead direction being the closed-form kernel
$\gamma^{-1}/\|\gamma^{-1}\|$ of the per-channel gain
\citep{shirodkar2026algebraicdeaddirectionslayernorm,TheoryRefNamed}. An architecture whose
degeneracy lives in an object none of these expose, an attention head's rotation or a dead expert
in a mixture, needs a detector built for that object, and supplying one extends the pipeline
without touching the read.

\paragraph{Duality-based identification.} A clean per-layer read needs the dead coordinate
at every layer, yet at each end of the network one K-FAC factor fails to pin it: at the
input $A_1=\mathrm{cov}(x)$ stays full rank, and at the output $G_L$ goes flat. The duality
$\lambda_{\min}(A_\ell)\,\lambda_{\min}(G_\ell)=\Theta(t^{2(L-1)})$ keeps the other factor
informative wherever one fails, so we read the dead coordinate from whichever factor
separates it. We fill the few layers neither factor resolves by mapping a neighbour's
coordinate through the layer Jacobian, then assemble the per-layer coordinates into the
cross-layer joint mode, the network-wide dead direction $u$ the read scans.

\paragraph{Construct, do not search.} The order-carrying joint mode sits above the bottom of
the Fisher spectrum. That bottom holds the gauge orbit, the $\sim(L-1)d^2$ within-layer
reparametrisation directions (for layer width $d$) that the layered structure leaves flat
\citep[Thm.~3]{TheoryRefNamed}, and below a sample ratio $n/d<1$ a sampling null space as
well; the order-carrying mode lies above both, since moving along it raises the Fisher as
$t^{2(k-1)}$ while the gauge orbit stays exactly flat. An eigendecomposition of the Fisher,
or any refinement toward its bottom, therefore returns a gauge or null direction instead of
the order, so we name the dead substructure from the factors and construct the joint mode
at it. The argument holds wherever the gauge orbit and the sampling null lie below the
order-carrying mode, which the gauge-quotient structure makes the generic case for
node-death.

\paragraph{Purity-matched rate window.} The order is the leading exponent of the rate as
$t\to0$, and the window selector recovers it without being given its value. A nominated
direction at cosine $1-\varepsilon$ to the true one carries an $\varepsilon$-sized component
on a lower-order direction, whose rate exponent lies $\Delta\alpha=2(k-k_{\mathrm{low}})$
below the true one. That component dominates the rate for
$t\lesssim\varepsilon^{2/\Delta\alpha}$, so the read is pure only above that scale.
Sweeping the lower end of the fit window once and keeping the highest-$r^2$ admissible
window ($r^2>0.95$) isolates the single power law that has cleared the contamination. The selector
optimizes the fit alone, so the architecturally fixed prediction then checks the recovered
order, having played no part in choosing the window. A robustness battery bears this out:
on planted orders the selector recovers $k\in\{2,3,4\}$ on every clean cell, returns no
order on a flat scan, and never matches a wrong exponent (Section~\ref{sec:taxonomy}).

\paragraph{What the read returns.} The scan sorts a direction by its slope and the magnitude it
reaches, over the whole axis between a flat gauge and a clean finite order. A genuine degeneracy
raises the directional Fisher as $t^{2(k-1)}$ with finite $k$, the slope returning the order. A
log-log slope $\alpha$ gives $k=1+\alpha/2$ and $\lambda_{\mathrm{dir}}=1/(2k)$: the squared-ReLU
unit of Figure~\ref{fig:rate_read}a rises with slope $4$, so $k=3$ and $\lambda_{\mathrm{dir}}=1/6$.
A flat scan is a gauge when it sits at a deep floor and a regular live direction when it sits at
the constant value a non-dead weight carries, a split the magnitude makes where the slope
cannot. The magnitude breaks a second tie: a curved gauge orbit read along its tangent rises
with slope $2$, the slope a $k{=}2$ node-death also shows, separated only by its deep floor.
Between the endpoints the slope itself diagnoses the direction. A slope shallower than the
prediction marks a contaminated direction, the dead mode mixed with a lower-order component
whose true exponent surfaces only above the contamination scale, the deviant per-coordinate
read of a rotated death, a returned value that misses the predicted order (Section~\ref{sec:taxonomy}). A slope steeper than the prediction marks
a generic line through a crossing of loci, the signal that routes the global assembly to its
structured resolution (Section~\ref{sec:discussion}). A slope that has not settled across the
window marks a pre-asymptotic scan. The read returns an order only on a clean, settled match
and otherwise rejects with the diagnosis, so it never mints a wrong exponent;
Appendix~\ref{app:diagnostic} is the key from slope and magnitude to verdict.

\paragraph{Reading the triple.} The read places three quantities in the frame of the Watanabe
triple $(\lambda, m, \nu)$, each with its own reach and status. From the order, each dead
direction carries the local coefficient $\lambda_{\mathrm{dir}}=1/(2k)$ exactly; the global
$\lambda$ assembles these by the sum-versus-crossing rule of Section~\ref{sec:background}
(independent directions add, a crossing takes the minimum) where the singular structure can be
enumerated, on the analytic models (Appendix~\ref{app:global}), and the posterior sampler
supplies it on a large network where the enumeration is open (Section~\ref{sec:comparison}). A
floor-aware count of the dead directions at the dominant order gives the dead-subspace
dimension; the constructed cells confirm it equals the analytic multiplicity $m$ under a normal
crossing of equal-order directions, while on a trained network it tracks the compressing dead
subspace (Section~\ref{sec:optimizer}). The singular fluctuation $\nu$ is fixed by the order, the
universal value $\nu(k)$, which a trained network's live structure suppresses below $\nu(k)$
(Section~\ref{sec:trajectory}).
 \section{A taxonomy of dead structure}
\label{sec:taxonomy}

The read is one instrument for a family of dead-structure types. Whatever the type, it
nominates a candidate direction from the factors, scans it, and classifies what it finds: a
genuine singularity carries a finite order the architecture fixes, a gauge stays at the floor.
Table~\ref{tab:taxonomy} maps the family and marks, for each type, whether a model in reach
formed a clean instance. The detected types carry the evidence below, the gauge family
completes at the transformer's architectural symmetries, and the attempted rows are types
whose precondition, a cleanly formed instance, the models tried did not meet.

A finite recovered order $k$ along a direction $u$ makes four claims at once. It certifies
that a genuine singularity has formed there: the directional Fisher rises from the floor with
a finite order, the signature of a dead direction the network drove its weights into. Its
value names the type: the activation's analytic order for a node-death, the depth for a
linear collapse, and no finite order for a gauge. It sets the local learning coefficient
$\lambda_{\mathrm{dir}}=1/(2k)$ this direction contributes to the model's complexity. And on the
constructed and architecture-known cells, where the order is set in advance, a match certifies
the read itself, so the same read can be trusted on a model whose singular structure is
unknown. The orders, dead-subspace dimensions, and gauge labels across a network then form a localized,
typed map of its singular complexity, the per-direction structure a single posterior-sampled
coefficient sums over but cannot localize.

\begin{figure}[t]
\centering
\includegraphics[width=0.56\textwidth]{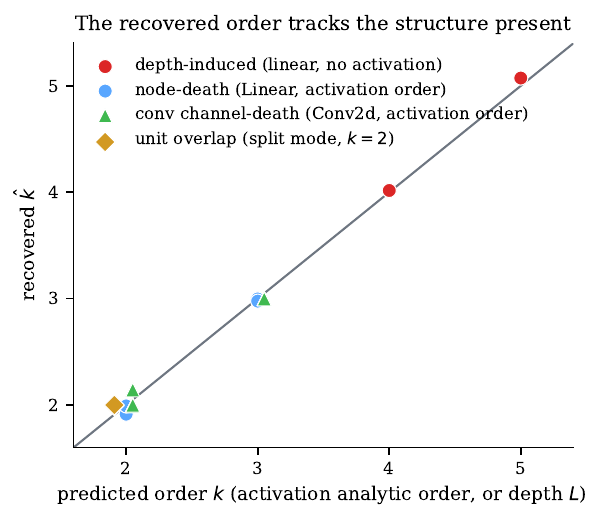}
\caption{The recovered order tracks the structure present. The predicted order is the
activation's analytic order for a node-death ($k{=}3$ squared-ReLU, $k{=}2$ gelu, across
$d{=}8$, $d{=}24$, and modular multiplication), the depth for a trained linear network
($k{=}L$, $L\in\{3,4,5\}$), and again the activation order for a convolutional channel read
through a different K-FAC factor. All fall on the diagonal.}
\label{fig:order}
\end{figure}

For the decomposition to mean anything, the recovered order must belong to the structure that
formed, fixed before the read. We check it where the order is predictable in advance, reading
structures whose $k$ we know and asking whether the scan returns it; across the genuine
singularities below the recovered order falls on the diagonal of the predicted one
(Figure~\ref{fig:order}).

\begin{table}[t]\centering\small
\caption{The family of dead-structure types the read covers. Order is the finite KL order a
genuine singularity carries, or a dash for a flat direction. Status marks the read state:
\emph{detected} here, or \emph{attempted}, scanned for but with no model in reach forming a
clean instance.}
\label{tab:taxonomy}
\begin{tabular}{lllll}
\toprule
type & kind & order & status & instance \\
\midrule
activation node-death       & singularity & activation order & detected & MLP, grokking, ViT \\
depth / determinantal       & singularity & $L$              & detected & deep linear network \\
convolutional channel-death & singularity & activation order & detected & CNN \\
unit overlap                & singularity & $2$ (curvature)  & detected & constructed \\
LayerNorm-kernel            & gauge       & --               & detected & DINOv2 \\
attention QK rotation       & gauge       & --               & detected & DINOv2 \\
ReLU rescaling              & gauge       & --               & detected & gauge companion \\
cross-entropy shift         & gauge       & --               & detected & constructed \\
attention VO rotation       & gauge       & --               & detected & DINOv2 \\
attention-head death        & singularity & head-level       & attempted & -- \\
mixture-of-experts expert   & input-dead  & --               & attempted & -- \\
\bottomrule
\end{tabular}
\end{table}

\subsection{Genuine singularities}

\paragraph{The activation sets a node-death's order.} The order of a node-death follows the
activation's local analytic order, through the order at which the activation derivative
$\phi'$ vanishes, so before
reading anything we can predict $k=2$ for gelu and ReLU and $k=3$ for squared-ReLU. To test
the prediction in isolation we construct a node-death in a small network, vary only the
activation, and scan the joint mode: the read returns $k=2,3,4$ for gelu, squared-ReLU, and
cubed-ReLU at $r^2=1.000$, recovering the predicted order in each case. To test it on a real
network we train a matched pair of grokking transformers \citep{PowerBurda22,NandaChanLieberum23} that differ only in activation and
read the dead structure each forms during training. The squared-ReLU model leaves its dead
subspace on the coordinate axes, where a scan along the axis returns $k=3$. The gelu model
leaves its dead subspace rotated off the axes, so a per-coordinate scan no longer follows
it and returns a deviant $1.29$; the off-canonical read identifies the rotated direction and
returns $k=2$ at $r^2=1.000$, the activation's predicted order recovered where the rotation
had hidden it. Appendix~\ref{app:constructed} gives the constructed node-death cells.

\paragraph{The same order on a real vision transformer.} Trained from
scratch on ImageNet under weight decay, a transformer forms the activation node-death the
fine-tuned model below lacks. A six-block ViT \citep{DosovitskiyBeiKolUszHou21} (width $256$, MLP hidden $1024$, patch $16$,
$112{\times}112$ inputs) on a $100$-class subset prunes its over-parametrised MLP as it
compresses, and the optimizer leaves the resulting dead subspace rotated off the coordinate
axes: at the deepest singular block the squared-ReLU coordinate-concentration is $0.07$. A
per-coordinate scan misses such a death; the off-canonical read recovers the
activation-predicted order on every condition (Table~\ref{tab:vit}), squared-ReLU
$\hat k = 3.00$ and gelu $\hat k \approx 2.0$, across two decay strengths and three seeds. On
a current vision architecture the order the read returns is the order the activation
fixes, recovered off a death the standard optimizer left rotated. Appendix~\ref{app:vit}
gives the setup for both vision-transformer regimes and the per-block gauge detection.

\begin{table}[ht]\centering\small
\caption{From-scratch ViT on ImageNet-100, per condition (mean over three seeds). The order
is read off the deepest singular block. Coordinate-concentration near $0$ is a dead subspace
rotated off the coordinate axes, near $1$ axis-aligned; the last column is the dead-subspace dimension (the Fisher-bottom count).}
\label{tab:vit}
\begin{tabular}{llllll}
\toprule
activation & wd & predicted $k$ & $\hat k$ & coord-conc. & dead-subsp. dim. \\
\midrule
squared-ReLU & $0.1$ & $3$ & $3.00$ & $0.07$ & $4$ \\
squared-ReLU & $0.2$ & $3$ & $3.00$ & $0.34$ & $2$ \\
gelu         & $0.1$ & $2$ & $2.02$ & $0.59$ & $14$ \\
gelu         & $0.2$ & $2$ & $2.03$ & $0.76$ & $100$ \\
\bottomrule
\end{tabular}
\end{table}

\paragraph{Depth sets a linear singularity's order.} A test that the read returns a
non-activation order needs a singularity with no activation behind it. We train a deep
linear network of depth $L$ to a rank-deficient target until one mode of the weight product
collapses to machine precision; the resulting dead direction takes its order from depth
alone, with $k=L$. Reading it at $L\in\{3,4,5\}$ returns $\hat k = 3.00, 4.02, 5.07$ at
$r^2=1.000$. The value $k=5$ is one no activation node-death produces, so the read is
tracking the order the structure carries. Appendix~\ref{app:constructed} gives the trained
deep-linear cells.

\paragraph{A convolutional channel carries the activation order through a different factor.}
The reads above run on the Linear MLP factor; a dead convolutional channel exposes its dead
direction through a spatial-patch covariance instead, a different K-FAC factor, which tests
whether the order law and the read survive the change of factor structure. We construct a
dead channel and read $k=2$ for ReLU and $k=3$ for squared-ReLU at $r^2=1.000$, and we then
train a wide CNN until weight decay kills its spare channels, nominate a dead channel from
the convolutional $G$-factor, and recover $k=2$ for ReLU and gelu and $k=3$ for squared-ReLU
across three seeds. The nominated channel-direction is usually a rotated combination of
channels, so the trained convolutional read is itself off canonical, and the activation
order survives both the new factor and the rotation. Appendix~\ref{app:conv} gives the
convolutional setup.

\paragraph{Task, width, and seed.} Three controls check that the node-death order is the
structure's, holding as task, width, and seed change. Changing the task to modular multiplication, the gelu
network still reads $k=2$ once it groks (off-canonical $\hat k=1.99$ at $r^2=1.000$).
Changing the width to $d{=}24$, three times the $d{=}8$ width and read at a proper sample
ratio $n/d\approx100$, the gauge-fixed squared-ReLU network reads $k=3$ along the axis and
the gauge-fixed gelu network $k=2$ off the axis, holding across three seeds and three
interior blocks ($\hat k = 2.81 \pm 0.18$ for squared-ReLU, $\hat k = 1.97 \pm 0.06$ for
gelu). A RoPE-aware attention gauge and a standard one give the same MLP order, so the
attention-side gauge choice leaves the read unchanged. Appendix~\ref{app:results} lists every per-cell read.

\paragraph{Two coincident units read the curvature order.} The other canonical two-layer
singularity is a unit overlap, two hidden units sharing the same incoming and outgoing
weights, the overlap singularity of \citet{AmariParkOzeki06} paired with the elimination
singularity a node-death realises. It carries two degenerate directions the read separates. Moving the outgoing
weights apart, $c_i\to c+s$ and $c_j\to c-s$, leaves the output unchanged, a flat transfer
gauge. Moving the incoming weights apart, $a_i\to a+t\delta$ and $a_j\to a-t\delta$, cancels
the odd terms of the expansion and grows the output as $t^2$ from the activation's curvature
$\phi''$ at the operating point, so the split direction carries order $k=2$. We construct an
overlap in a small network and read $k=2$ on the split at $r^2=1.000$ for squared-ReLU and
gelu alike, the transfer direction flat at the floor. The split order is the curvature's,
$k=2$ for either activation, while a node-death's order tracks the activation's order at zero;
so one architecture carries two singularity types whose orders the read tells apart. For $n$
coincident units the split order stays $k=2$ and its multiplicity grows as $n-1$, the transfer
gauge of the same dimension, which we read at $n=2,3,4$. Appendix~\ref{app:constructed} gives
the overlap construction.

\subsection{Architectural gauges}

\paragraph{A real model often carries a gauge in place of a node-death.} A DINOv2 ViT-S
\citep{oquab2023dinov2} fine-tuned on CIFAR-100 carries
no weight-space node-death: scanning every transformer block's feed-forward layer for a
hidden unit whose incoming and outgoing weights have both collapsed, the smallest combined
norm stays at $0.44$ to $0.73$ of the block median across all twelve blocks, with no unit
near the floor. The dead structure this model carries is a gauge, which the read classifies
directly. Appendix~\ref{app:vit} gives the fine-tuned DINOv2 setup and the per-block gauge reads.

\paragraph{The LayerNorm-kernel gauge.} What this model carries
instead is the LayerNorm-kernel direction $u^\star=\gamma^{-1}/\|\gamma^{-1}\|$
\citep{shirodkar2026algebraicdeaddirectionslayernorm,TheoryRefNamed}, the exact kernel of the post-LayerNorm input
covariance: at every block the detected $u^\star$ aligns with the covariance's
smallest-eigenvalue direction to $|\cos|=1.0000$. The read flags it as a gauge, its
directional Fisher sitting $10^4$ to $5\times10^{4}$ below a live direction at the two
normalisation sites of each block. The flatness comes from the LayerNorm centering
symmetry. Projecting the activations onto $u^\star$ gives a variance at machine precision
($\mathrm{std}=8\times10^{-16}$) about a nonzero mean set by the LayerNorm bias, so along
$u^\star$ the data is effectively constant. The direction is therefore an approximate gauge,
and the residual Fisher is that constant offset carried through the nonlinearity, a small
nonzero floor.

\paragraph{The attention rotation gauge, and the slope it imitates.} The attention query--key
rotation is
a second architectural gauge: a shared per-head rotation of the query and key projections
leaves every attention score invariant, and once the rotation includes the projection bias
the read places it $10^{6}$ below a live direction. This gauge is a curved orbit, so read
along its tangent it rises with slope $2$, the slope a $k{=}2$ node-death also shows, and
only the depth of its floor separates the two. The read therefore classifies the
architectural gauges of this network as flat and reserves a finite order for genuine
node-death, the distinction the magnitude makes (Section~\ref{sec:method}).

\paragraph{Completing the gauge family.} The LayerNorm scale and the query--key rotation are
two of the transformer's architectural symmetries, the gauge family the equivariant optimizer
of \citet{DDCRef} is built to quotient; the same flat-magnitude test reaches the rest. The attention value--output rotation is the sibling of the query--key gauge under the
per-head $O(d_{\mathrm{head}})$ symmetry. The cross-entropy shift is a constant added to every
output logit that the softmax absorbs. The ReLU rescaling is the single-sided move at a hidden
unit, the gauge companion the order reads use throughout. Each is a flat direction with no
order, a contributor to the architectural multiplicity. We read the ReLU rescaling directly as
that companion, and a constructed cell reads the cross-entropy shift flat, its directional
Fisher more than twenty orders of magnitude below a live bias direction. The value--output
rotation reads the same way on the fine-tuned vision transformer, a curved gauge rising with
slope $2$ but sitting four orders of magnitude below a live direction.

\subsection{Reach and open testbeds}

The read mechanism, nominate then scan then classify, does not change with the type, so the
family extends past the cells above once a clean instance forms. Attention-head death forms in reach as a low
weight-space rank of the attention block (Appendix~\ref{app:component_death}), with no single
head carrying an order; an isolated head-death needs a known-prunable-head transformer. A
mixture-of-experts expert goes dead by distribution coverage, reviving under a different corpus,
and a forced-dead expert is input-dead, reading flat with no order; its durable weight-level
structure is the within-expert node-death already in the table. A clean instance of either forms
in models we did not read here, so we leave the full validation to future work; that formed
instance is the precondition the read shares with every type it does detect.
 \section{Optimizer and training phase shape the dead structure}
\label{sec:optimizer}

The activation and the depth fix the order, and the optimizer and the training phase fix
the basis the dead structure occupies and whether it forms at all.

\paragraph{The orthogonaliser decides whether a clean structure forms.} The optimizer
decides whether a cleanly readable dead structure forms at all, and on this deep transformer
the orthogonaliser is what decides it, with the gauge projection playing the separate role we
isolate below. We read the same run
under three optimizers from one matched cohort over three seeds: vanilla Muon \citep{jordan2024muon} (the
textbook degree-five Newton--Schulz orthogonalisation, NS5, the \texttt{ns\_off} baseline), the
scaled-polar orthogonaliser with the gauge removed (the
\textsc{DDCMuon} orthogonaliser of \citealp{DDCRef}, \texttt{bf\_fast}), and the
gauge-equivariant optimizer on that orthogonaliser (Appendix~\ref{app:basis}). Vanilla Muon
leaves the dead structure diffuse, a small flat near-kernel spread across directions with its
axis-alignment falling from $0.71$ to $0.40$ through depth and no single direction carrying a
clean order; the scan reports a deviant pre-asymptotic value under both activations
($\hat k = 1.13 \pm 0.15$ for gelu and $1.48 \pm 0.43$ for squared-ReLU, $r^2$ below the
admissibility threshold), so the read finds no order to recover. The scaled-polar
orthogonaliser, with the gauge removed, instead compresses the network into a large clean
dead subspace: under squared-ReLU its directions sit on the coordinate axes at alignment
$>0.99$ and read $k=3$, and under gelu they sit rotated off the axes, where the off-canonical
read recovers $k=2$. The gauge-equivariant optimizer on the same orthogonaliser reads the
same order at the same alignment, so the readable axis-aligned basis comes from the
scaled-polar orthogonaliser, and the gauge's separate effect is to spectrally separate the
bottom block. The contrast reproduces at the task-natural one-block width, where weight decay
matched across the arms confirms the orthogonaliser supplies the alignment
(Appendix~\ref{app:basis}).

\begin{figure}[t]
\centering
\includegraphics[width=0.62\textwidth]{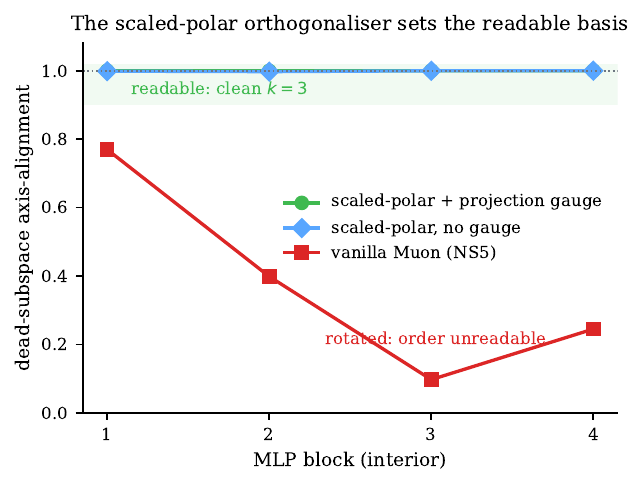}
\caption{The orthogonaliser sets the basis. Dead-subspace axis-alignment by interior block
for three optimizers from one matched cohort: the scaled-polar orthogonaliser, with or
without the gauge projection, keeps the subspace on the coordinate axes (alignment
$\approx1$, clean $k{=}3$), while vanilla Muon rotates it off the axes with falling alignment
through depth, where a per-coordinate read cannot follow it. The readable axis-aligned basis
comes from the orthogonaliser; the gauge arm reads the same.}
\label{fig:optimizer}
\end{figure}

The off-canonical read removes the canonical-alignment precondition, recovering a dead
direction whether it sits on the coordinate axes or rotated off them, as the gelu case
shows. It keeps the one remaining precondition: the frozen checkpoint must sit at the
singularity in the nominated direction, so the synthetic scan out from it shows the clean
$t^{2(k-1)}$ growth the admissibility gate checks. The optimizers reach different
checkpoints. The scaled-polar orthogonaliser lands the units at clean node-deaths, read at
$k=3$ on the axes or $k=2$ rotated; vanilla Muon, on this deep transformer, lands
at a diffuse, low-alignment checkpoint whose best nominated direction stays pre-asymptotic
($r^2$ below the gate). That is a property of the
solution the optimizer reaches: the common optimizers on the fixed architecture below reach
readable structures with no gauge fixing (Section~\ref{sec:discussion}). Under the per-coordinate squared-ReLU the gauge-fixed
rotation is no symmetry, so the clean rotated subspace is geometry the optimizer produced
(Figure~\ref{fig:optimizer}).

\paragraph{The common optimizers on a fixed architecture.} The reads above move the
optimizer and the depth together. To place the optimizer on its own we hold a small
architecture fixed, a deterministic squared-ReLU teacher-student MLP whose spare units a
wide student prunes to a node-death under weight decay, and read it under SGD, AdamW
\citep{LoshchilovHutter19}, RMSProp \citep{TielemanHinton12}, and Adam \citep{KingmaBa15} at three seeds (Table~\ref{tab:optaxis}; the cell setup is
Appendix~\ref{app:optaxis}). The target is deterministic, so
the spare units have nothing to fit and prune cleanly under every optimizer; this removes the
adaptive-preconditioner resistance a noisy target provokes, where the Adam family amplifies
the spare units' noise-fitting gradient. The off-canonical rate read
returns $k\approx3$ (asymptotic, $r^2=1.0$) under every optimizer; the optimizer sets the
structure the order occupies. SGD, which carries no preconditioner to bend the metric, and
Adam leave a clean node-death on the coordinate axes, where the gauge companion (a
single-sided move at the dead unit) reads flat. AdamW leaves a deep node-death rotated off
the axes at one of the three seeds, where the off-canonical read supplies the order a
per-coordinate scan would miss. RMSProp, a pure diagonal preconditioner without momentum,
keeps a distributed representation with no unit driven to a weight-space joint-zero, so the
rate reads $k\approx3$ off the lowest-Fisher direction while the gauge companion stays
finite, an order read off a near-kernel with no confirmed node-death. On the ReLU cells SGD follows
the activation order ($k\approx2$), and the single-seed AdamW ReLU cell reads deviant. Across
the common optimizers on this deterministic architecture the read recovers the order wherever
a low-Fisher direction forms. The noisy target is now measured (Table~\ref{tab:noisytarget}). The deep grokking transformer
under vanilla Muon keeps its structure diffuse, and no clean dead direction forms, so the
admissibility gate correctly returns no order there, the null reading that marks the absence of
a formed singularity.

\begin{table}[t]
\centering
\caption{Optimiser-axis coverage on a fixed squared-ReLU teacher-student node-death cell
(predicted $k=3$, three seeds). The off-canonical order reads $k\approx3$ under every
optimizer; the optimizer sets the structure the order occupies. The gauge companion is a
single-sided move at the dead unit, flat only when the consumer has been pruned (a
confirmed node-death).}
\label{tab:optaxis}
\begin{tabular}{lccll}
\toprule
optimizer & $\hat k$ & $r^2$ & axis-alignment & structure (gauge companion) \\
\midrule
SGD     & $3.00$ & $1.00$ & $0.58$--$1.00$ & clean / rotated node-death (flat) \\
Adam    & $3.00$ & $1.00$ & $1.00$         & clean axis node-death (flat) \\
AdamW   & $2.99$ & $1.00$ & $0.72$--$1.00$ & deep, sometimes rotated (flat) \\
RMSProp & $2.98$ & $1.00$ & $0.58$--$0.79$ & distributed near-kernel (finite) \\
\bottomrule
\end{tabular}
\end{table}

\paragraph{A noisy target grades the precondition by optimizer.} The deterministic cell
prunes the spare units cleanly under every optimizer. Gaussian noise on the teacher target
gives the adaptive preconditioner a spare-unit noise-fitting gradient to amplify, and the
prune then separates the optimizers (Table~\ref{tab:noisytarget}, sweeping the noise scale
$\sigma_{\mathrm{train}}$ across $\{0, 0.1, 0.25, 0.5, 1.0\}$ at three seeds, the read-time
geometry on the true-MC Fisher). SGD forms a clean $k{=}3$ node-death at every noise level,
the death deepening as the noise grows. Adam under its coupled-$L_2$ weight decay holds the
clean axis node-death across the sweep. RMSProp keeps its distributed near-kernel, the rate
reading $k\approx3$ off the lowest-Fisher direction with no confirmed node-death. AdamW breaks
at the first noise level: its spare units fit the noise, no unit reaches a joint-zero, and the
order read returns the deviant value. The data-fit shows in the empirical-over-true-MC
bottom-eigenvalue ratio, which grows to $94$ for AdamW and $100$ for RMSProp as the noise
rises, the inflation the true-MC geometry sees through. Where a clean death forms (SGD, Adam),
the true-MC bottom eigenvalue sits below the residual's resolution, leaving the ratio
undefined. The off-canonical read follows the geometry under target noise wherever a
low-Fisher direction survives, and returns the deviant value where the optimizer fits the
noise and no unit prunes.

\begin{table}[t]
\centering
\caption{Noisy-target sweep on the same squared-ReLU node-death cell (predicted $k=3$, three
seeds), the target gaining Gaussian noise of scale $\sigma_{\mathrm{train}}$ and the read-time
geometry on the true-MC Fisher. The prune separates the optimizers the deterministic cell
prunes alike: SGD and Adam hold the clean death, RMSProp keeps a near-kernel, AdamW loses the
death once the target is noisy. The data-fit ratio (empirical over true-MC bottom eigenvalue)
is undefined where a clean death forms, its true-MC eigenvalue sitting below the residual's
resolution.}
\label{tab:noisytarget}
\begin{tabular}{lccc}
\toprule
optimizer & at $\sigma_{\mathrm{train}}{=}0$ & at $\sigma_{\mathrm{train}}{\ge}0.1$ & data-fit ratio \\
\midrule
SGD     & clean $k{=}3$ death       & clean $k{=}3$ death          & (death below floor) \\
Adam    & clean $k{=}3$ death       & clean $k{=}3$ death (mostly) & (death below floor) \\
RMSProp & near-kernel $k{\approx}3$ & near-kernel $k{\approx}3$    & $1.9 \to 100$ \\
AdamW   & clean $k{=}3$ death       & order deviant                & $0.9 \to 94$ \\
\bottomrule
\end{tabular}
\end{table}

\paragraph{The dead-subspace dimension.} Counting the dead directions at a block gives the
dead-subspace dimension, the per-direction read supplying the order and a floor-aware count
supplying how many directions share it; under a normal crossing of equal-order directions this
count is the analytic multiplicity. The count keeps the genuine node-deaths and drops the gauge directions
and the below-floor numerical near-kernels, so the architectural symmetries (the
$\sim(L{-}1)d^2$ reparametrisation directions and the per-site LayerNorm-centering
directions) enter a separate architectural multiplicity, distinct from the learned one.

\paragraph{The structure emerges in the compression phase.} Read across training steps, the
dead subspace grows and sharpens: its dimension rises from $188$ to $562$ while its
axis-alignment climbs from $0.78$ to $0.999$, with the order $k=3$ steady throughout
(Appendix~\ref{app:developmental}). The
gauge directions are present from initialisation, fixed by the architecture, whereas the
node-deaths appear only as the network compresses, which gives the classification a second,
developmental signature beyond the frozen-checkpoint read. In the accumulation
regime, before the network compresses, no node-death forms and the read returns no order, a
regime boundary the discussion makes precise (Section~\ref{sec:discussion}).

 \section{The singular fluctuation: fixed by the order, absorbed by the network}
\label{sec:trajectory}

The third invariant of the Watanabe triple, the singular fluctuation $\nu$, sets
$\mathrm{WAIC}=T_n+2\nu/n$ and the asymptotic train--validation gap. Along a dead direction $\nu$
is fixed by the order alone: it is a universal function $\nu(k)$ of the KL order, independent of
the model-specific leading coefficient \citep{TheoryRefNamed}, with $\nu(2)\approx0.173$ and
$\nu(3)\approx0.278$. The order read reaches $\nu$ the way it reaches $\lambda_{\mathrm{dir}}$:
from $k$. The reach has a ceiling we measure here, where the live structure of a trained network
absorbs part of the fluctuation.

\paragraph{The order fixes $\nu(k)$ on an isolated direction.} We confirm the predicted $\nu(k)$
by measuring it directly on data, an independent check of a value the theory computes by
quadrature. On the canonical order-$k$ cell
$y=a\,s^{k}+\varepsilon$, $\varepsilon\sim\mathcal{N}(0,\sigma^{2})$, we form the exact posterior
on generated noisy data and compute Watanabe's functional variance
$V_n=\sum_i \mathrm{Var}_{\mathrm{post}}[\log p(y_i\mid s)]$, with $\hat\nu=V_n/2$. Data-averaged
over $300$ draws, $\hat\nu$ recovers $\nu(k)$ within a few percent across
$n\in\{500,\dots,4000\}$: $0.167$ to $0.178$ at $k{=}2$ (against $0.173$) and $0.275$ to $0.289$
at $k{=}3$ (against $0.278$). The functional-variance estimator is itself calibrated against the
regular anchor, where it returns $\nu=d/2$ on a $d$-parameter linear-Gaussian model by both the
closed form and a posterior sampler (the cells are Appendix~\ref{app:nu}).

\begin{figure}[t]
\centering
\includegraphics[width=\textwidth]{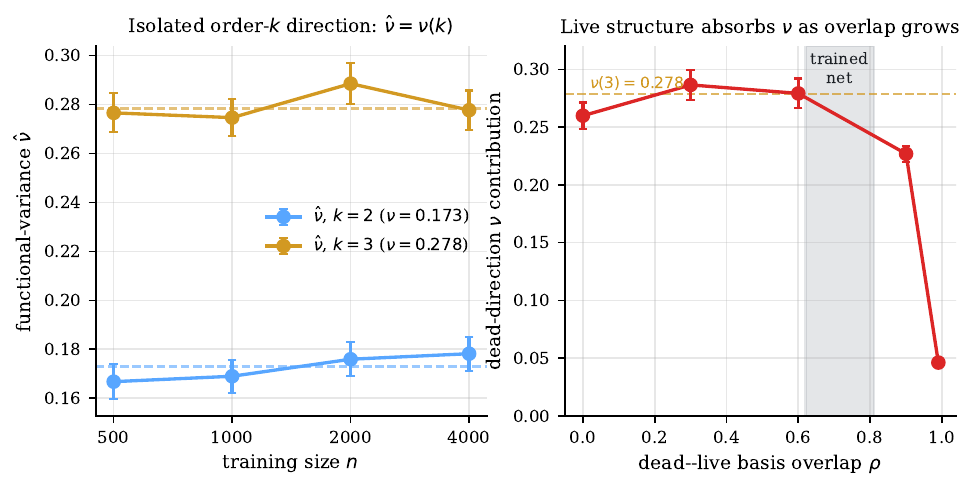}
\caption{The order fixes $\nu(k)$, the live structure absorbs it. Left: on the isolated order-$k$
cell $y=a\,s^{k}+\varepsilon$ the functional-variance $\hat\nu$ recovers the universality value
$\nu(k)$ across training size $n$, data-averaged over $300$ draws ($\nu(2){=}0.173$,
$\nu(3){=}0.278$). Right: in the controlled cell $y=b\,g+s^{k}c+\varepsilon$, the dead
direction's contribution to $\nu$ holds at $\nu(3)$ while the regular and dead bases stay
distinct and collapses as their overlap $\rho=\mathrm{corr}(g,c)$ grows ($120$ draws per point);
the shaded band is the effective dead--live overlap $\rho_{\mathrm{eff}}\in[0.62,0.81]$ measured
in the trained vision transformers, where the live structure absorbs the fluctuation.}
\label{fig:nu-absorption}
\end{figure}

\paragraph{A trained network's realized $\nu$ falls below $\nu(k)$.} The universality value is the
fluctuation an isolated order-$k$ direction sees. In a trained network the dead direction's basis
overlaps the live units, and the live parameters absorb part of the data fluctuation the
singular fluctuation integrates over. We isolate the effect on a controlled cell
$y=b\,g(x)+s^{k}c(x)+\varepsilon$ with one regular parameter $b$ (basis $g$) and one order-$k$
dead coordinate $s$ (basis $c$): in the joint, data-averaged posterior the dead direction
contributes exactly $\nu(k)$ while the bases stay distinct, and the contribution collapses as
they align (Figure~\ref{fig:nu-absorption}). In the trained over-parametrised networks of
Section~\ref{sec:taxonomy} the effective overlap of a generic dead-scan basis with the live-unit
span is $\rho_{\mathrm{eff}}\approx0.62$ to $0.81$, inside the suppression band, so the live
structure absorbs the fluctuation and the realized $\nu$ sits below $\nu(k)$ by a
structure-dependent amount. The order fixes the idealised $\nu(k)$; recovering a trained
network's realized $\nu$ needs the live-structure absorption, which the order alone does not
carry.

\paragraph{The frozen scan is the calibrated order.} The order the read reports comes from the
frozen scan, synthesized at the checkpoint. A dead direction also collapses along the training
trajectory, but its $\sigma_{\min}$ decay against steps is the optimizer's approach speed, which
on the grokking runs separates $k=3$ from $k=2$ at close to chance. The trajectory carries the
order in the Fisher-against-$\sigma_{\min}$ slope under a theorem-compatible descent, validated
under SGD \citep{shirodkar2026deaddirectionsignaturescheapspectral} and scoped out for the
Adam-class and Muon runs here. The frozen scan removes that descent precondition; the trajectory
collapse then cross-checks the frozen order without supplying it.
 \section{The global view: alongside posterior sampling}
\label{sec:comparison}

The global view estimates the single threshold by sampling the local posterior with
SGLD \citep{LauFurmanWangMurfetWei25}, and setting that estimate beside the geometry read
fixes what each measures and how its accuracy is established.

\paragraph{Accuracy is fixed on closed-form ground truth.} On the analytic models that anchor
accuracy (Section~\ref{sec:background}) the geometry read recovers the per-direction order
exactly, $k\in\{2,3,4\}$ on the planted node-deaths and $k=L$ on the deep linear networks,
while the sampler's local coefficient carries its documented width-dependent drift from the
global threshold and is read as a rank statistic.

\paragraph{A scalar against a decomposition.} The sampler returns one global number and
localises at most to a parameter subset, a layer or an attention head, its refined form
\citep{WangHoogland24}; a single direction lies below its resolution. The per-direction order,
the intersection type, and the gauge-versus-singularity verdict have no counterpart in a single
scalar, so the paper validates them against closed forms; the sampler has no per-direction read
to set beside them. The geometry read returns the order of each dead direction and
the dead-subspace dimension of the directions sharing it, the per-direction decomposition the global
scalar aggregates by the
sum-versus-crossing rule of Section~\ref{sec:background}. The two answer different
questions: the geometry read assembles the global coefficient where the singular structure is
enumerable, on the analytic models, and the sampler supplies it where enumeration on a large
network is still open (Section~\ref{sec:discussion}).

\paragraph{Cost.} The geometry read is deterministic and needs no calibration, returning
the order and the dead-subspace dimension in one pass of forward and backward evaluations. A credible
sampler estimate needs a per-model calibration sweep to a stability plateau before the
estimate, so the cost gap is the calibration overhead and the per-model retuning that
precede it (Table~\ref{tab:llc}; the full head-to-head is Appendix~\ref{app:llc}).

\paragraph{The triple, assembled.} From the geometry view we read the per-direction order and
the dead-subspace dimension, and through the order the universal $\nu(k)$ (live-basis absorption
suppresses the realized value in a trained network). The global view adds the single threshold
the sampler estimates. The first two reads need no canonical alignment and no posterior sample,
and they decompose the sampler's global value.
 \section{Uses}
\label{sec:uses}

Beyond a single complexity scalar the read returns two things: a per-direction decomposition
into the order and the dead-subspace dimension, and a gauge-versus-singularity verdict. Several uses follow from what the paper
already measures, and more are immediate given how cheap the read is.

\paragraph{A gauge-fixing diagnostic.} The classification names which
directions a gauge-equivariant optimizer \citep{DDCRef} should quotient away and which carry
learned structure to keep, a flat verdict marking a symmetry direction and a finite order a
node-death. On the fine-tuned vision transformer the read returns this verdict directly
(Section~\ref{sec:taxonomy}), turning the optimizer's gauge choices from a design
assumption into a measured target.

\paragraph{The effective dimension tracks the sampler.} The cheap read that
matches the sampler most closely is the curvature-weighted effective dimension of the K-FAC
Fisher spectrum, $\hat\lambda_{\mathrm{eff}}(\gamma)=\tfrac12\sum_i\lambda_i/(\lambda_i+\gamma)$
over the block's factor-eigenvalue products, at one forward and backward pass per block. On a
width-128 seven-optimizer cohort it recovers the calibrated SGLD coefficient restricted to each
block at Spearman $\rho=0.82$ on the MLP and $\rho=0.79$ on the attention
(Appendix~\ref{app:ranking_perblock}). The curvature weighting is what carries it: a bare rank or
dead-unit count does not track ($\rho=0.29$ on this cohort), since the sampler weights each
direction by its curvature. The liveness gate reads a fully input-dead MLP to zero, the
coefficient the restricted sampler also reads. The decomposition is the informative object,
since the global coefficient is near-constant across these arms and the per-block restriction
carries the cross-optimizer structure.

\begin{figure}[t]
\centering
\includegraphics[width=0.58\textwidth]{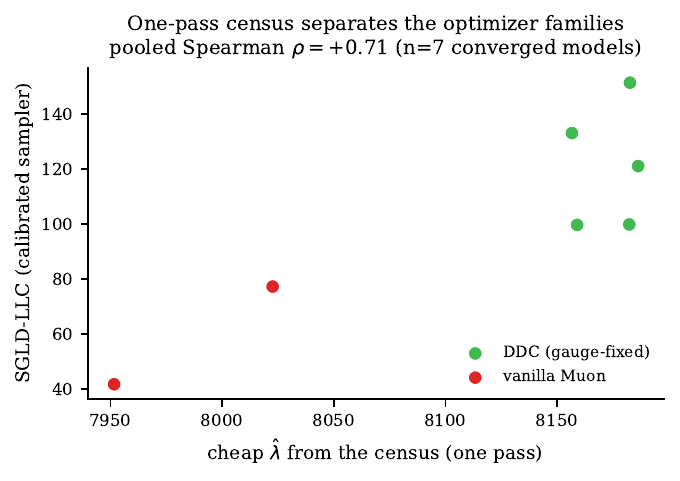}
\caption{The one-pass census separates the optimizer families. The census learning-coefficient
estimate $\hat\lambda$ (one forward pass) against the calibrated SGLD-LLC (a per-model
calibration sweep), on the d8 modular-addition models where the sampler converges.
Across the seven distinct models every gauge-fixed DDC model (green) reads as more complex than
every vanilla-Muon model (red) in both estimators; the pooled rank agreement is the weaker
Spearman $\rho = 0.71$ across those models and $\rho = 0.66$ over the eleven checkpoints.}
\label{fig:complexity_ranking}
\end{figure}

\paragraph{A one-pass cross-optimizer separator.} The dead-unit census of
\citet{shirodkar2026deaddirectionsignaturescheapspectral} is cheaper still, a single forward
pass with no spectrum. Its estimate $\hat\lambda = N_{\mathrm{total}}/2 - N_{\mathrm{dead}}(1/2 -
1/(2k))$ is the regular dimension minus the reduction the dead units buy. On eighteen d8
modular-addition checkpoints (gauge-fixed DDC, vanilla Muon, vanilla AdamW, with and without
weight decay), across the seven distinct models where the sampler converges ($\hat R\le1.1$,
positive estimate), every gauge-fixed DDC model reads as more complex than every vanilla-Muon
model in both the census and the SGLD-LLC (Figure~\ref{fig:complexity_ranking}). The separator is
the dead-unit count, which the census turns into $\hat\lambda$ directly for this order-$k{=}3$
set: $17$ to $106$ dead directions in the DDC models against $508$ to $721$ in the vanilla-Muon
ones. That count is only $0.1\%$ to $4.4\%$ of the $N_{\mathrm{total}}$ regular dimension, so
$\hat\lambda$ barely moves off $N_{\mathrm{total}}/2$: it separates the families cleanly but not
the checkpoints within a run, where the count follows the weight-decay schedule while the sampler
moves the other way, and the pooled agreement is the weaker Spearman $\rho=0.71$ across the seven
models and $\rho=0.66$ over the eleven checkpoints. Where the sampler degenerates to a negative
estimate, on the vanilla-AdamW and accumulation-regime cells, the census still returns a reading.
Appendix~\ref{app:ranking} gives the full ranking.

\paragraph{Developmental tracking.}
The read is cheap enough to run at every checkpoint, tracing the singular structure through
training: the gauge directions are present from initialisation and the node-deaths emerge as
the network compresses (Appendix~\ref{app:developmental}). The per-model profile (its order
spectrum, gauge count, and effective Fisher rank by layer) is a richer signature than the
bottom singular value alone. The per-direction diagnosis sharpens it into model-level
measurements: a compression-progress fraction from the settled-versus-in-transit split, a
continuous deadness-depth spectrum in place of a binary count, and a bucket-census fingerprint.
Read alongside a run, the same profile is a training diagnostic: it shows whether the singular
structure is forming, separating a compressing network from one the optimizer leaves diffuse
(Section~\ref{sec:optimizer}) or one still in the accumulation regime with no node-death, a
geometric state the loss curve does not report. The read is cheap enough to monitor across the
run as it trains.
 \section{Discussion and limitations}
\label{sec:discussion}

This paper turns the order-recovery of \citet{TheoryRefNamed} into a measurement: a
descent-free read at a frozen checkpoint that decomposes the single learning-coefficient
scalar into per-direction orders and a dead-subspace dimension, classifies each direction as a genuine
singularity or a flat gauge, and does so off canonical alignment across transformer,
convolutional, and normalisation layers. What follows maps where the read applies, what it
needs to run, and the cases at the programme's frontier.

\paragraph{The read applies once a singularity has formed.} The read needs a dead direction
in place, which a network forms in the compression phase as it folds into a singular
solution. A checkpoint still accumulating capacity, where the bottom singular value rises,
carries no dead structure, and the read correctly returns no order: singular-learning analysis
begins once a singularity has formed. Standard language-model pretraining sits in the
accumulation phase, so the read characterises its compressed checkpoints and reports the
absence of singular structure on the rest.

\paragraph{A detector has to surface the structure.} The read runs on whatever a detector
exposes, and the paper supplies detectors across the dead-structure family: the K-FAC
factors, the convolutional channel factor, the LayerNorm kernel, the attention rotations, and
the constructed overlap and shift cells (Table~\ref{tab:taxonomy}). Each returns a classified
verdict, a finite order where a node-death or depth singularity has formed and a gauge
contribution where the structure is a symmetry, both of which the read reports as results.
Two rows stay unverified, attention-head death and expert death in a mixture of experts: a
detector is in hand for each, but no model in reach formed a clean instance to read, so they
sit at attempted in the table. The detector family is finite, so the completeness of an
enumeration is itself measurable: the K-FAC factors flag every separated dead block per layer
regardless of type, and comparing that total against the dimension the named detectors claimed
returns the unattributed remainder, dead structure the factors see but no detector typed. A
nonzero remainder bounds what the named reads miss, so a global coefficient assembled from the
named structure alone is a lower bound on the complexity until the remainder is closed.

\paragraph{What the read tells you on an arbitrary model.} The read reads dead structure it is
pointed at; it does not find structure on its own. A uniformly random direction in a random
network almost surely reads regular, since the order-carrying directions are a constructed,
measure-zero set and the order-carrying mode sits above the high-dimensional gauge floor, so a
random line has vanishing overlap with it. Every order or gauge verdict is therefore
conditional on a detector having nominated the direction, and searching the parameter space at
random or refining the Fisher toward its bottom returns a gauge or null direction in place of
the order (Section~\ref{sec:method}). Given any model and any weights the read does three
things accurately. It certifies a detector-nominated direction, returning a finite order with
$\lambda_{\mathrm{dir}}=1/(2k)$ or a gauge with none, and rejecting a contaminated, crossing, or
unsettled scan with a diagnosis and never a wrong order. It reports the absence of singular
structure on an uncompressed checkpoint, a real answer about the network's phase. And it
decomposes a known, enumerable structure into per-direction orders the analytic models confirm
to machine precision. The hard limit is the first sentence: the read sees the fraction of a
real model's singular structure its detector family surfaces (the unattributed remainder is
the coverage residual above). The bottom singular value, the K-FAC bottom-block separation, and
the deadness ratio answer, before any scan, whether a checkpoint has compressed enough to carry
anything to read.

\paragraph{The optimizer shapes how clean the read is.} The read characterises whatever
structure an optimizer leaves, and different optimizers leave different solutions
(Section~\ref{sec:optimizer}). The architecture fixes the order; the basis and the sharpness
are the optimizer's. An orthogonalising optimizer \citep{DDCRef} leaves an axis-clean structure
that reads directly, a standard one can leave it rotated, which the off-canonical read still
follows, and on a deep transformer it can leave the structure too diffuse to carry any order.
The read sees only what the optimizer leaves; where that is nothing, as for vanilla Muon on the
deep grokking transformer, there is nothing to recover.

\paragraph{$\sigma_{\min}$ holds only in regime.} The bottom singular value underwrites the
magnitude floor and the pre-scan compression check. At deeply singular checkpoints its magnitude
becomes unreliable, nearing the precision floor as its sample variance grows, and the
asymptoticity and floor gates reject the read once it reaches the floor. The rank-level reading,
the effective rank and the cross-checkpoint ordering, still holds at those checkpoints; the
absolute magnitude is what is lost.

\paragraph{The global coefficient.} The per-direction reads assemble into a global learning
coefficient by the sum-versus-crossing rule of Section~\ref{sec:background} wherever the
singular structure is enumerable. Across twelve such cells, the normal crossings, separable
sums, their composites, and the scalar deep linear network at depths two through five, the
assembled $(\lambda, m)$ reproduces the closed-form threshold to machine precision
(Table~\ref{tab:global}); the scalar network's depth-$L$ collapse reads order $k=L$, and the
global $\lambda=1/2$ follows from the $L$-hyperplane crossing. The blind assembly breaks on
one structure, the wide matrix deep linear networks, whose product-zero locus is
determinantal with its crossings rotated off every coordinate. There the coordinate read
mis-counts and the assembled coefficient leaves the closed form, a gap the improved rotated
order read does not close, since it sits in the assembly. Three reads carry across the
boundary, developed in Appendix~\ref{app:global}: a rigorous bracket from the
Newton-polyhedron simplex bound and the generic-line order contains the closed form on every
cell and holds the hundreds-scale $\lambda$ of the trained $H=(20,h,h,20)$ network up to
$h=128$; a blind detector flags the determinantal locus and recovers its depth; and the
structured resolution rule returns the exact coefficient on all five determinantal cells for
depth $L\le3$. A general exact resolver for an arbitrary determinantal variety stays open,
the simple origin blow-up stalling at the simplex bound, an item in the programme's
open-problems register. For one global number there the posterior sampler stays the practical
tool, with the geometry read supplying the per-direction decomposition where individual
directions can be targeted. The multiplicity recovers the way the coefficient does: typing the
dominant-order directions returns it where they form a single crossing, robust to padding by
lower-order, higher-order, and gauge directions, and under-counts only when the multiplicity
splits across several equal-threshold loci, which then need enumerating. The singular
fluctuation reaches a sharper limit: the order fixes the universal $\nu(k)$, but a trained
network's live structure absorbs part of the data fluctuation, so the realized $\nu$ sits
below $\nu(k)$ by a structure-dependent amount the order alone does not carry
(Section~\ref{sec:trajectory}).

\paragraph{Cost and sample count.} The proxies the read uses cost little: the activation
$\sigma_{\min}$ is one SVD, the K-FAC factor reads are per-layer, and the directional Fisher
is a few forward evaluations; only a full-spectrum Fisher would grow cubically in width, and
the read does not form one. The geometry reads carry an $n/d$ sample-count requirement, met
at the $d=24$ width ($n/d\approx100$) and approached at $d=8$ ($n/d\approx64$), where the
recovered orders still hold at $r^2=1.000$.
 
\clearpage
\bibliographystyle{plainnat}

\begin{thebibliography}{30}
\providecommand{\natexlab}[1]{#1}
\providecommand{\url}[1]{\texttt{#1}}
\expandafter\ifx\csname urlstyle\endcsname\relax
  \providecommand{\doi}[1]{doi: #1}\else
  \providecommand{\doi}{doi: \begingroup \urlstyle{rm}\Url}\fi

\bibitem[Amari(2016)]{Amari16}
Shun-ichi Amari.
\newblock \emph{Information Geometry and Its Applications}, volume 194 of
  \emph{Applied Mathematical Sciences}.
\newblock Springer, 2016.
\newblock URL \url{https://link.springer.com/book/10.1007/978-4-431-55978-8}.

\bibitem[Amari et~al.(2006)Amari, Park, and Ozeki]{AmariParkOzeki06}
Shun-ichi Amari, Hyeyoung Park, and Tomoko Ozeki.
\newblock Singularities affect dynamics of learning in neuromanifolds.
\newblock \emph{Neural Computation}, 18\penalty0 (5):\penalty0 1007--1065,
  2006.
\newblock URL \url{https://doi.org/10.1162/neco.2006.18.5.1007}.

\bibitem[Aoyagi(2024)]{Aoyagi24}
Miki Aoyagi.
\newblock Consideration on the learning efficiency of multiple-layered neural
  networks with linear units.
\newblock \emph{Neural Networks}, 172:\penalty0 106132, 2024.
\newblock URL \url{https://doi.org/10.1016/j.neunet.2024.106132}.

\bibitem[Aoyagi and Watanabe(2005)]{AoyagiWatanabe05}
Miki Aoyagi and Sumio Watanabe.
\newblock Stochastic complexities of reduced rank regression in {B}ayesian
  estimation.
\newblock \emph{Neural Networks}, 18\penalty0 (7):\penalty0 924--933, 2005.
\newblock URL \url{https://doi.org/10.1016/j.neunet.2005.03.014}.

\bibitem[Bushnaq et~al.(2024)Bushnaq, Mendel, Heimersheim, Braun,
  Goldowsky-Dill, H\"anni, Wu, and Hobbhahn]{BushnaqMendel24}
Lucius Bushnaq, Jake Mendel, Stefan Heimersheim, Dan Braun, Nicholas
  Goldowsky-Dill, Kaarel H\"anni, Cindy Wu, and Marius Hobbhahn.
\newblock Using degeneracy in the loss landscape for mechanistic
  interpretability, 2024.
\newblock URL \url{https://arxiv.org/abs/2405.10927}.

\bibitem[Carroll(2021)]{Carroll21}
Liam Carroll.
\newblock Phase transitions in neural networks.
\newblock Master's thesis, School of Mathematics and Statistics, The University
  of Melbourne, 2021.
\newblock URL \url{http://therisingsea.org/notes/MSc-Carroll.pdf}.

\bibitem[Dosovitskiy et~al.(2021)Dosovitskiy, Beyer, Kolesnikov, Weissenborn,
  Zhai, Unterthiner, Dehghani, Minderer, Heigold, Gelly, Uszkoreit, and
  Houlsby]{DosovitskiyBeiKolUszHou21}
Alexey Dosovitskiy, Lucas Beyer, Alexander Kolesnikov, Dirk Weissenborn,
  Xiaohua Zhai, Thomas Unterthiner, Mostafa Dehghani, Matthias Minderer, Georg
  Heigold, Sylvain Gelly, Jakob Uszkoreit, and Neil Houlsby.
\newblock An image is worth 16x16 words: Transformers for image recognition at
  scale.
\newblock In \emph{ICLR}, 2021.

\bibitem[Farrugia-Roberts(2022)]{FarrugiaRoberts22}
Matthew Farrugia-Roberts.
\newblock Structural degeneracy in neural networks.
\newblock Master's thesis, School of Computing and Information Systems, The
  University of Melbourne, 2022.
\newblock URL \url{https://far.in.net/mthesis}.

\bibitem[Farrugia-Roberts(2023)]{FarrugiaRoberts23}
Matthew Farrugia-Roberts.
\newblock Functional equivalence and path connectivity of reducible hyperbolic
  tangent networks.
\newblock In \emph{Advances in Neural Information Processing Systems 36
  (NeurIPS)}, pages 79502--79517, 2023.
\newblock URL \url{https://arxiv.org/abs/2305.05089}.

\bibitem[Grosse and Martens(2016)]{GrosseMartens16}
Roger Grosse and James Martens.
\newblock A {K}ronecker-factored approximate {F}isher matrix for convolution
  layers.
\newblock In \emph{ICML}, 2016.
\newblock URL \url{https://arxiv.org/abs/1602.01407}.

\bibitem[Hironaka(1964)]{Hironaka64}
Heisuke Hironaka.
\newblock Resolution of singularities of an algebraic variety over a field of
  characteristic zero.
\newblock \emph{Annals of Mathematics}, 79\penalty0 (1):\penalty0 109--326,
  1964.
\newblock URL \url{https://www.jstor.org/stable/1970486}.

\bibitem[Hoogland et~al.(2024)Hoogland, Wang, Farrugia-Roberts, Carroll, Wei,
  and Murfet]{HooglandWangFarrugiaRoberts24}
Jesse Hoogland, George Wang, Matthew Farrugia-Roberts, Liam Carroll, Susan Wei,
  and Daniel Murfet.
\newblock Loss landscape degeneracy and stagewise development in transformers.
\newblock \emph{Transactions on Machine Learning Research}, 2024.
\newblock URL \url{https://arxiv.org/abs/2402.02364}.

\bibitem[Jordan et~al.(2024)]{jordan2024muon}
Keller Jordan et~al.
\newblock Muon: An optimizer for hidden layers in neural networks.
\newblock \url{https://kellerjordan.github.io/posts/muon/}, 2024.
\newblock Blog post.

\bibitem[Kingma and Ba(2015)]{KingmaBa15}
Diederik~P. Kingma and Jimmy Ba.
\newblock Adam: A method for stochastic optimization.
\newblock In \emph{International Conference on Learning Representations
  (ICLR)}, 2015.
\newblock URL \url{https://arxiv.org/abs/1412.6980}.

\bibitem[Lau et~al.(2025)Lau, Furman, Wang, Murfet, and
  Wei]{LauFurmanWangMurfetWei25}
Edmund Lau, Zach Furman, George Wang, Daniel Murfet, and Susan Wei.
\newblock The local learning coefficient: A singularity-aware complexity
  measure.
\newblock In \emph{AISTATS}, 2025.
\newblock URL \url{https://proceedings.mlr.press/v258/lau25a.html}.

\bibitem[Loshchilov and Hutter(2019)]{LoshchilovHutter19}
Ilya Loshchilov and Frank Hutter.
\newblock Decoupled weight decay regularization.
\newblock In \emph{International Conference on Learning Representations
  (ICLR)}, 2019.
\newblock URL \url{https://arxiv.org/abs/1711.05101}.

\bibitem[Martens and Grosse(2015)]{MartensGrosse15}
James Martens and Roger Grosse.
\newblock Optimizing neural networks with {Kronecker}-factored approximate
  curvature.
\newblock In \emph{ICML}, 2015.
\newblock URL \url{https://arxiv.org/abs/1503.05671}.

\bibitem[Nanda et~al.(2023)Nanda, Chan, Lieberum, Smith, and
  Steinhardt]{NandaChanLieberum23}
Neel Nanda, Lawrence Chan, Tom Lieberum, Jess Smith, and Jacob Steinhardt.
\newblock Progress measures for grokking via mechanistic interpretability.
\newblock In \emph{ICLR}, 2023.
\newblock URL \url{https://arxiv.org/abs/2301.05217}.

\bibitem[Oquab et~al.(2023)Oquab, Darcet, Moutakanni, Vo, Szafraniec, Khalidov,
  Fernandez, Haziza, Massa, El-Nouby, et~al.]{oquab2023dinov2}
Maxime Oquab, Timoth\'{e}e Darcet, Th\'{e}o Moutakanni, Huy Vo, Marc
  Szafraniec, Vasil Khalidov, Pierre Fernandez, Daniel Haziza, Francisco Massa,
  Alaaeldin El-Nouby, et~al.
\newblock {DINOv2}: Learning robust visual features without supervision.
\newblock \emph{arXiv preprint arXiv:2304.07193}, 2023.

\bibitem[Power et~al.(2022)Power, Burda, Edwards, Babuschkin, and
  Misra]{PowerBurda22}
Alethea Power, Yuri Burda, Harri Edwards, Igor Babuschkin, and Vedant Misra.
\newblock Grokking: Generalization beyond overfitting on small algorithmic
  datasets.
\newblock \emph{arXiv:2201.02177}, 2022.

\bibitem[Roy and Vetterli(2007)]{RoyVetterli07}
Olivier Roy and Martin Vetterli.
\newblock The effective rank: A measure of effective dimensionality.
\newblock \emph{15th European Signal Processing Conference (EUSIPCO)}, pages
  606--610, 2007.

\bibitem[Shirodkar(2026{\natexlab{a}})]{DDCRef}
Tejas~Pradeep Shirodkar.
\newblock {Dead-Direction Conditioners: Gauge-Equivariant Preconditioning for
  Deep Networks}, 2026{\natexlab{a}}.
\newblock URL \url{https://arxiv.org/abs/2606.29176}.

\bibitem[Shirodkar(2026{\natexlab{b}})]{TheoryRefNamed}
Tejas~Pradeep Shirodkar.
\newblock Dead directions: Geometric singular learning, 2026{\natexlab{b}}.
\newblock URL \url{https://arxiv.org/abs/2606.05957}.

\bibitem[Shirodkar and
  Narayanan(2026{\natexlab{a}})]{shirodkar2026algebraicdeaddirectionslayernorm}
Tejas~Pradeep Shirodkar and P.~J. Narayanan.
\newblock Algebraic dead directions in {LayerNorm} transformers: A
  forward-pass-only diagnostic at {LLM} scale, 2026{\natexlab{a}}.
\newblock URL \url{https://arxiv.org/abs/2606.19491}.

\bibitem[Shirodkar and
  Narayanan(2026{\natexlab{b}})]{shirodkar2026deaddirectionsignaturescheapspectral}
Tejas~Pradeep Shirodkar and P.~J. Narayanan.
\newblock Dead-direction signatures: A cheap spectral reading of singular
  complexity, 2026{\natexlab{b}}.
\newblock URL \url{https://arxiv.org/abs/2606.21158}.

\bibitem[Tieleman and Hinton(2012)]{TielemanHinton12}
Tijmen Tieleman and Geoffrey Hinton.
\newblock Lecture 6.5---{RMSProp}: Divide the gradient by a running average of
  its recent magnitude.
\newblock COURSERA: Neural Networks for Machine Learning, 2012.

\bibitem[Timaeus and collaborators(2024)]{devinterp}
Timaeus and collaborators.
\newblock {devinterp}: A library for developmental interpretability.
\newblock \url{https://github.com/timaeus-research/devinterp}, 2024.
\newblock Python package.

\bibitem[Wang et~al.(2025)Wang, Hoogland, van Wingerden, Furman, and
  Murfet]{WangHoogland24}
George Wang, Jesse Hoogland, Stan van Wingerden, Zach Furman, and Daniel
  Murfet.
\newblock Differentiation and specialization of attention heads via the refined
  local learning coefficient.
\newblock In \emph{International Conference on Learning Representations
  (ICLR)}, 2025.
\newblock URL \url{https://arxiv.org/abs/2410.02984}.
\newblock Spotlight.

\bibitem[Watanabe(2009)]{Watanabe09}
Sumio Watanabe.
\newblock \emph{Algebraic Geometry and Statistical Learning Theory}.
\newblock Cambridge University Press, 2009.
\newblock URL \url{https://doi.org/10.1017/CBO9780511800474}.

\bibitem[Watanabe(2018)]{Watanabe18}
Sumio Watanabe.
\newblock \emph{Mathematical Theory of {B}ayesian Statistics}.
\newblock CRC Press, 2018.
\newblock URL \url{https://www.routledge.com/9781482238068}.

\end{thebibliography}

\clearpage
\appendix
\section*{Appendices}
\pdfbookmark[0]{Appendices}{appendix-root}

We give the experiments in full here, from the network setups through to the
global-coefficient assembly and the singular-fluctuation cells.

\etocsettocdepth.toc{subsection}
{\renewcommand{\contentsname}{Appendix contents}\tableofcontents}
\bigskip

\section{The read: setup and pipeline}\label{app:sec:read}

\subsection{Experimental setup}
\label{app:setup}
The paper reads four families of network. The modular-addition transformers are one-block RoPE
(rotary position embedding) models trained to predict $(a+b)\bmod p$ from the two operand
tokens, read at the grokked checkpoint where the network has folded into its singular solution.
All d=8 runs use $5000$ steps and the d=24 run $6000$, at seed $42$ unless noted. The
gauge-fixed optimizer is Muon with a body-frame rotation gauge on the attention QK and VO blocks
(the \textsc{DDCMuon} recipe of \citealp{DDCRef}); vanilla Muon is the same recipe with the
gauge removed (textbook Newton--Schulz, \texttt{ns\_off}). The deep linear networks are trained
by gradient descent to a rank-deficient regression target until one mode of the weight product
reaches machine precision, with a hand-constructed linear bridge swept toward its singularity
for method validation. The convolutional cell is a teacher--student channel-death network
(Appendix~\ref{app:conv}). The two vision transformers, a DINOv2 ViT-S/14 fine-tuned on
CIFAR-100 and a from-scratch six-block ViT trained on an ImageNet-100 subset, are set up with
their reads in Appendix~\ref{app:vit}. Table~\ref{tab:cohorts} lists the transformer, linear,
and convolutional cohorts.

The transformer reads accumulate the K-FAC factors over held-out tokens and scan the true
(label-resampled) softmax-categorical Fisher (Appendix~\ref{app:pipeline}), at sample ratio
$n/d\approx64$ for the d=8 reads and $n/d\approx100$ for the d=24 reads. The frozen-checkpoint
scans run in single precision (float32), since lower precision floors the directional Fisher
before the power-law window opens. Each read and figure regenerates from its committed driver
and result JSON.

\begin{table}[ht]\centering\small
\caption{Cohorts. Width $d$ is the model dimension; the MLP hidden width is $4d$.}
\label{tab:cohorts}
\setlength{\tabcolsep}{4pt}
\begin{tabular}{l c p{3.0cm} p{2.5cm} p{3.4cm}}
\toprule
cohort & $d$ & optimizer & activation & role \\
\midrule
base            & 8  & gauge-fixed, vanilla (3 seeds) & squared-ReLU & unit, multiplicity, depth, optimizer \\
activation pair & 8  & gauge-fixed & squared-ReLU, gelu & activation order \\
multiplication  & 8  & gauge-fixed & squared-ReLU, gelu & task generality (modular mult.) \\
ReLU            & 8  & gauge-fixed & relu & standard activation \\
width           & 24 & gauge-fixed, vanilla & squared-ReLU, gelu & model width \\
deep linear     & 6  & gradient descent & none (linear) & non-activation (depth) order \\
linear bridge   & 6  & constructed & none (linear) & method validation \\
conv CNN        & --  & construct; Adam$+$WD (teacher--student) & relu, squared-ReLU, gelu & conv channel-death (factor structure) \\
\bottomrule
\end{tabular}
\end{table}

\subsection{The read pipeline}
\label{app:pipeline}
The read takes a checkpoint and a target MLP block. It accumulates the K-FAC
factors $A$ and $G$ over held-out tokens, nominates the dead direction from the
better-separated factor (the A--G duality), and assembles the cross-layer joint
mode. It then scans the directional Fisher along that mode, $\theta(t)=\theta_0+t
u$, evaluating $u^\top \fisher(\theta(t)) u$ by a finite-difference
Jacobian-vector product on the true (label-resampled) Fisher, and fits the log-log
slope over the auto-selected purity window. The sample ratio $n/d$ is set by the
number of accumulation batches: $n/d\approx64$ for the d=8 reads and $n/d\approx100$
for the d=24 reads. The scan grid is $16$ points over $t\in[10^{-2},0.5]$. The
window selector keeps the highest-$r^2$ admissible single-power-law fit.

The read needs no library beyond a forward pass and one singular value decomposition. Given a
nominated unit-norm direction $u$ at a base point $\theta_0$, it scans the true
softmax-categorical Fisher quadratic form $u^\top \fisher(\theta)u$ along
$\theta(t)=\theta_0+tu$ by a finite-difference Jacobian-vector product, fits a single power
law over the purity window, and classifies the result.
\begin{small}
\begin{verbatim}
# inputs: model, batch x; unit direction u at base theta0; finite-diff step eps;
#         log-spaced scan grid t_values (16 pts in [1e-2, 0.5]).

def dir_fisher(theta, u):     # u^T F u, softmax-CE Fisher (residual-free)
    f  = lambda th: logits(model_at(th), x)   # shifted-param logits
    Ju = (f(theta + eps*u) - f(theta - eps*u)) / (2*eps)
    p  = softmax(f(theta))                    # (N tokens, C classes)
    return mean_x[ sum_c p*Ju^2 - (sum_c p*Ju)^2 ]   # cat-cov contraction

ufu       = [dir_fisher(theta0 + t*u, u) for t in t_values]   # the scan
alpha, r2 = best_powerlaw_window(t_values, ufu)   # sweep lower cut, keep max-r2
k_hat     = 1 + alpha/2

# classify (verdict = asymptotic iff r2 > 0.95 and the window clears the floor):
if asymptotic and r2>0.95 and k_hat>=1.5:  genuine singularity (order k_hat)
elif max(ufu) << a live direction:         gauge (no finite order)
else:                                      live / pre-asymptotic
\end{verbatim}
\end{small}
At an exact constructed singularity $u^\top \fisher(\theta_0)u=0$ for a genuine direction and
a gauge alike, so the clean rise of the scan, not the base-point magnitude, separates them; a
flat gauge never rises, a curved gauge rises with slope $2$ but from a deep floor, and a
genuine degeneracy rises from a live-scale coefficient.

\subsection{A diagnostic key}
\label{app:diagnostic}
The read returns four numbers per direction: the slope $\alpha$ (equivalently the order
$\hat k = 1+\alpha/2$), the single-power-law fit quality $r^2$, the asymptoticity verdict
(whether the slope has settled across the window), and the magnitude $u^\top\fisher u$ the scan
reaches relative to a live direction. Table~\ref{tab:diagnostic} is the key from these four to
the verdict and to what it says about the nominated direction. The classification is
conservative: it accepts an order only on a clean, settled match (the first row) and otherwise
rejects with a named diagnosis, so it never returns a wrong exponent. The four reject rows are
the failure modes the off-canonical read is built to tell apart, and each names its remedy: a
shallow slope is a contaminated direction to re-nominate, a steep one a crossing to resolve
structurally, an unsettled one a scan to widen, and a flat live one a direction that is simply
not dead.

\begin{table}[ht]\centering\small
\caption{Reading the scan: slope and magnitude to verdict. ``deep floor'' is a magnitude orders
of magnitude below a live direction's Fisher; ``live'' is the live-scale value a non-dead weight
carries; $k$ is the predicted order where one is known (the activation order for a node-death,
the depth for a determinantal collapse).}
\label{tab:diagnostic}
\setlength{\tabcolsep}{4pt}
\begin{tabular}{l l l l p{4.9cm}}
\toprule
$\hat k$ & magnitude & settled & verdict & the nominated direction \\
\midrule
$\approx k$       & live       & yes    & singularity   & a genuine degeneracy of order $k$, returning $\lambda_{\mathrm{dir}}=1/(2k)$ \\
$\approx 1$       & deep floor & --     & gauge         & a flat symmetry, a multiplicity contributor with no order \\
$\approx 2$       & deep floor & yes    & gauge         & a curved gauge orbit imitating $k{=}2$, separated by its floor and not its slope \\
$\approx 1$       & $10^{-3}$--$10^{-1}$ live & -- & near-dead & depressed but not floored, ambiguously dead; the continuous depth $\log_{10}(F_{\max}/F_{\mathrm{live}})$ resolves it \\
$\approx 1$       & live       & --     & regular       & not dead ($k{=}1$), a live weight direction \\
$1<\hat k<k$      & live       & either & impure        & contaminated, the dead mode mixed with a lower-order component, re-nominate \\
$\hat k>k$        & live       & yes    & composite     & a generic line through a crossing of loci, route to the structured resolution \\
any rise          & --         & no     & pre-asymptotic & the slope has not settled, scan wider or the structure has not compressed \\
\bottomrule
\end{tabular}
\end{table}

Each verdict carries a refinement the read also returns. The magnitude is reported as the
continuous depth $\log_{10}(F_{\max}/F_{\mathrm{live}})$, so the gauge / near-dead / regular
split reads as a position on the depth continuum. The slope is reported as a
profile, a plateau for a clean power law, a monotone drift for a transitional structure, or a
curved log-log, from the local slopes the single window discards. A pre-asymptotic reject is
split in turn: a slope drifting up from a depressed magnitude is a unit still compressing
(\emph{in transit}), while a flat-slope one only needs a wider scan. These read directly off
the scan the verdict already computed.

Table~\ref{tab:real-reads} applies the key to existing checkpoint reads. The grokking
node-deaths return their order, on the axis and recovered off it; the two DINOv2 gauges read
flat at a deep floor, the query--key rotation rising with slope $2$ but separated from a
$k{=}2$ node-death by its $10^{-6}$ floor; and the diffuse vanilla-Muon read returns no
settled order. The classifications come from the diagnostic key applied to the committed
read data, with no re-run.

\begin{table}[ht]\centering\small
\caption{The diagnostic key on real-model reads. Each row is an existing checkpoint read
classified by the diagnostic key (Table~\ref{tab:diagnostic}); \emph{depth} is
$\log_{10}(F/F_{\mathrm{live}})$ where a live reference is recorded.}
\label{tab:real-reads}
\setlength{\tabcolsep}{4pt}
\begin{tabular}{l l l l l l}
\toprule
read & $\hat k$ & $r^2$ & slope & depth & classification \\
\midrule
grokking node-death (squared-ReLU) & $2.99$ & $1.00$ & asymptotic & --   & singularity \\
grokking gelu, off-canonical       & $2.07$ & $1.00$ & asymptotic & --   & singularity \\
DINOv2 LayerNorm-kernel            & $1.00$ & --     & flat       & $-4$ & gauge \\
DINOv2 QK rotation                 & $2.00$ & --     & slope $2$  & $-6$ & gauge \\
vanilla Muon, rotated read         & $1.59$ & $0.71$ & unsettled  & --   & pre-asymptotic \\
\bottomrule
\end{tabular}
\end{table}

 \section{Per-type reads}\label{app:sec:pertype}

\subsection{Per-cell results}
\label{app:results}
\begin{figure}[ht]
\centering
\includegraphics[width=0.62\textwidth]{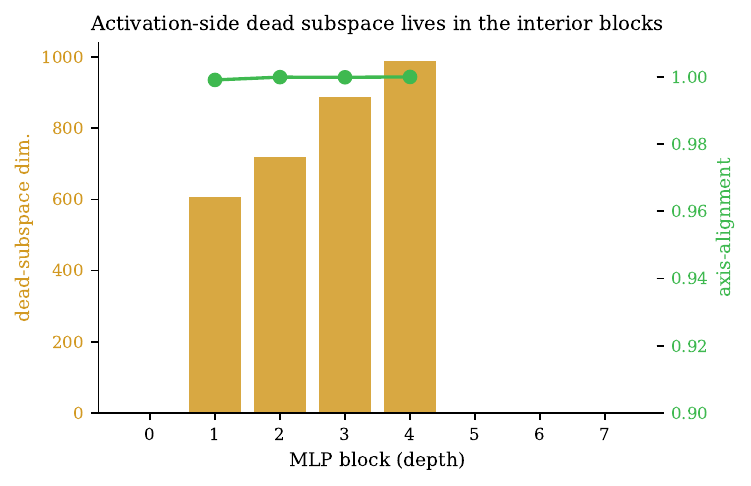}
\caption{Dead subspace by depth (gauge-fixed d=8). The axis-aligned
activation-side subspace sits in the interior blocks ($1$--$4$) and grows with
depth; the input block carries its dead structure on the gradient factor (A-side
$m{=}0$), and the late blocks carry none.}
\label{fig:depth}
\end{figure}

Table~\ref{tab:results} reports the read per cell. The cells fall into the three verdicts the
method returns. The gauge-fixed squared-ReLU blocks read a clean node-death, $\hat k\approx3$
at axis-alignment above $0.99$ and $r^2=1.000$, the order the activation fixes. The gelu cells
read $\hat k\approx2$ at axis-alignment near $0.1$, the same order recovered off the
coordinate axes through the rotated-direction read, where a per-coordinate scan finds nothing.
The vanilla-Muon base reads pre-asymptotic, its dead structure diffuse with no single
direction carrying a clean order ($r^2=0.71$, no order returned). The trained deep linear
cells read the depth order $\hat k=L$ for $L=3,4,5$ with no activation behind them, $k=5$ an
order no node-death produces. Architecture fixes the order, the optimizer the alignment, and the read separates the two.
Figure~\ref{fig:depth} shows
the dead-subspace dimension and axis-alignment by depth for the gauge-fixed d=8 run.

\begin{table}[ht]\centering\small
\caption{Per-cell reads. Axis-alignment near $1$ means the dead subspace lies on the
coordinate axes; $\hat k$ is the recovered order (single-axis where axis-aligned,
rotated-direction read where not).}
\label{tab:results}
\setlength{\tabcolsep}{4pt}
\begin{tabular}{p{2.7cm} l l l l l p{1.9cm}}
\toprule
cohort / cell & block & $m$ & axis-align & $\hat k$ & $r^2$ & verdict \\
\midrule
base, gauge                   & 1  & 608 & 0.999 & 3.00 & 1.000 & asymptotic \\
base, vanilla                 & 1  & 123 & 0.71  & deviant & 0.71 & pre-asymptotic \\
activation pair, squared-ReLU & 1  & 715 & 0.998 & 2.99 & 1.000 & asymptotic \\
activation pair, gelu         & 1  & 59  & 0.09  & 1.99 & 1.000 & asymptotic (rotated) \\
multiplication, gelu          & 1  & --  & 0.10  & 1.99 & 1.000 & asymptotic (rotated) \\
width, squared-ReLU           & 12 & --  & --    & 2.98 & 0.999 & asymptotic \\
width, gelu                   & 12 & --  & --    & 1.99 & 1.000 & asymptotic (rotated) \\
deep linear $L{=}3,4,5$       & -- & 1   & rotated & 3.00, 4.02, 5.07 & 1.000 & asymptotic \\
\bottomrule
\end{tabular}
\end{table}

\subsection{Constructed and trained linear cells}
\label{app:constructed}
The constructed and trained-linear cells here read a structure whose order is fixed before
the measurement, which checks that the read returns the order planted before the scan.

\paragraph{Constructed node-death (planted order).} In a small two-layer network one
hidden unit's incoming and outgoing weights are zeroed, the activation is varied, and
the joint mode is scanned. The read returns the activation's analytic order, $k=2,3,4$
for gelu, squared-ReLU, and cubed-ReLU at $r^2=1.000$ (Figure~\ref{fig:order}). The same
construction on a convolutional channel (Appendix~\ref{app:conv}) reads the order
through the spatial-patch covariance, a different K-FAC factor. The selector recovers
the planted order on every clean cell, returns no order on a flat scan, and never
matches a wrong exponent.

\paragraph{Trained deep linear network (depth order).} A deep linear network of depth
$L$ is trained by gradient descent to a rank-deficient regression target until one mode
of the weight product collapses to machine precision. The resulting dead direction takes
its order from depth alone, $k=L$, with no activation behind it. Reading it at
$L\in\{3,4,5\}$ returns $\hat k=3.00,4.02,5.07$ at $r^2=1.000$ (Table~\ref{tab:results}),
and $k=5$ is an order no activation node-death produces. The wide matrix deep linear
networks, whose singular set is a determinantal variety, mark the boundary of the global
assembly and are treated in Appendix~\ref{app:global}.

\paragraph{Per-type direction constructors.} Each type plants its degeneracy and supplies the
directions $u$ the read scans, so the constructed-cell claims reproduce from the description
alone. Write $a_j$ for hidden unit $j$'s incoming weights, $c_j$ for its outgoing weights, and
$d,d'$ for random unit vectors.
\begin{small}
\begin{verbatim}
node-death (unit j):  set a_j = c_j = 0;  theta0 = 0 in the (a_j, c_j) block
          joint mode u = [d ; d'] / sqrt(2)
          -> order k = activation order (sq-ReLU 3, gelu/ReLU 2)
unit overlap (units 0..n-1 all copied onto unit 0 so they coincide):
          split_i    = +d  on a_0, -d  on a_i  -> order k = 2 (curvature phi'')
          transfer_i = +d' on c_0, -d' on c_i  -> flat gauge  (i = 1..n-1)
cross-entropy shift (output bias b in R^C):  u = 1_C / sqrt(C)  -> flat gauge
ReLU rescale (units W_l, W_{l+1}): one-sided log-norm move      -> flat gauge
\end{verbatim}
\end{small}
The LayerNorm-kernel gauge is read in closed form, the kernel
$u^\star=\gamma^{-1}/\|\gamma^{-1}\|$ of the per-channel gain $\gamma$, with no scan. The
attention rotation gauges are curved: for an antisymmetric generator $A$ on a head's
$d_{\mathrm{head}}$ subspace, the query--key gauge moves $\delta W_Q = W_Q A$,
$\delta W_K = -A W_K$, and the value--output gauge moves $\delta W_V = W_V A$,
$\delta W_O = -A W_O$; both leave the head invariant, so the read places them at a deep floor
(the magnitude criterion, since they rise with slope $2$ along the tangent).

\subsection{Convolutional channel-death}
\label{app:conv}
The convolutional read uses a small CNN: two $3{\times}3$ convolutions
($c_{\mathrm{in}}{=}3 \to 12 \to 12$, padding $1$, no bias), the activation, a
global average pool, and a linear head to a $4$-dimensional regression target. The
second convolution carries the dead channel; the head is the channel's consumer.
Inputs are $512$ random $3{\times}8{\times}8$ images. In the constructed case one
channel's incoming filter and outgoing head slice are zeroed, and the scan runs
along the joint turn-on mode. In the trained case the target is a narrow teacher CNN
($4$ channels) the wide student over-covers, trained with Adam (lr $3{\cdot}10^{-3}$,
weight decay $5{\cdot}10^{-2}$, $4000$ steps); the spare channels die, and the dead
channel-direction is nominated from the smallest-eigenvalue eigenvector of the
convolutional $G$-factor (the $C_{\mathrm{out}}{\times}C_{\mathrm{out}}$
per-spatial-position output-gradient covariance). The frozen-weight true-MC Fisher
scan and the purity window are the same as the Linear reads. The nominated direction
is usually a rotated combination of channels (axis-alignment near $0$ for the smooth
and polynomial activations), so the trained read is off canonical.

\begin{table}[ht]\centering\small
\caption{Convolutional channel-death reads. Trained rows are the mean recovered
order over three seeds $\{42,142,242\}$.}
\label{tab:conv}
\begin{tabular}{lllll}
\toprule
case & activation & predicted $k$ & $\hat k$ & $r^2$ \\
\midrule
construct          & relu          & 2 & 2.00 & 1.000 \\
construct          & squared-ReLU  & 3 & 3.00 & 1.000 \\
construct          & gelu          & 2 & 2.14 & 1.000 \\
trained (3 seeds)  & relu          & 2 & 2.00 & 1.000 \\
trained (3 seeds)  & squared-ReLU  & 3 & 3.00 & 1.000 \\
trained (3 seeds)  & gelu          & 2 & 2.00 & 1.000 \\
\bottomrule
\end{tabular}
\end{table}

\subsection{Vision transformer reads}
\label{app:vit}
The real vision-transformer reads of Section~\ref{sec:taxonomy} cover two regimes on
the same architecture: a fine-tuned network that carries a gauge, and a from-scratch
network that forms a node-death.

\paragraph{Setup.} The fine-tuned model is a DINOv2 ViT-S/14 (twelve blocks, embedding
dimension $384$, patch $14$) fine-tuned on CIFAR-100, read at the converged checkpoint.
The from-scratch model is a six-block ViT (width $256$, MLP hidden $1024$, patch $16$,
$112\times112$ inputs) trained on a $100$-class ImageNet subset under weight decay, read
at the deepest singular block. The detector at each block is the closed-form
LayerNorm-kernel direction $u^\star=\gamma^{-1}/\|\gamma^{-1}\|$ for the gauge read and
the smallest-eigenvalue eigenvector of the squared-ReLU input covariance for the
node-death read; the directional Fisher scan and the purity window are the same as the
Linear reads.

\paragraph{The fine-tuned model carries gauges.} Across all twelve
blocks the smallest combined feed-forward norm sits at $0.44$ to $0.73$ of the block
median, so no weight-space node-death has formed. What the model carries is three
architectural gauges, all read as flat (Table~\ref{tab:vit-gauge}): the LayerNorm
kernel, whose detected $u^\star$ aligns with the input covariance's smallest-eigenvalue
direction at $|\cos|=1.0000$ and whose directional Fisher sits orders of magnitude below
a live direction, and the attention query-key and value-output rotations, read $10^6$ and
$2\times10^4$ below live with the slope-$2$ a curved gauge orbit imitates. The read
reserves a finite order for a genuine node-death and flags these as gauges.

\begin{table}[ht]\centering\small
\caption{Architectural gauges on the fine-tuned DINOv2 ViT-S/14. The LayerNorm-kernel
direction is detected at three blocks; the QK and VO rotations at block $6$. The gauge/live
ratio is the directional Fisher against a live direction at the same site; a deep floor
marks a gauge.}
\label{tab:vit-gauge}
\begin{tabular}{lllll}
\toprule
gauge & block & $|\cos(u^\star,\text{min-eigvec})|$ & gauge/live Fisher & slope \\
\midrule
LayerNorm kernel & $0$  & $1.0000$ & $4\times10^{-5}$ & flat \\
LayerNorm kernel & $6$  & $1.0000$ & $1\times10^{-4}$ & flat \\
LayerNorm kernel & $11$ & $1.0000$ & $2\times10^{-2}$ & flat \\
QK rotation      & $6$  & --       & $1\times10^{-6}$ & $2.00$ \\
VO rotation      & $6$  & --       & $5\times10^{-5}$ & $2.00$ \\
\bottomrule
\end{tabular}
\end{table}

\begin{figure}[ht]
\centering
\includegraphics[width=0.78\textwidth]{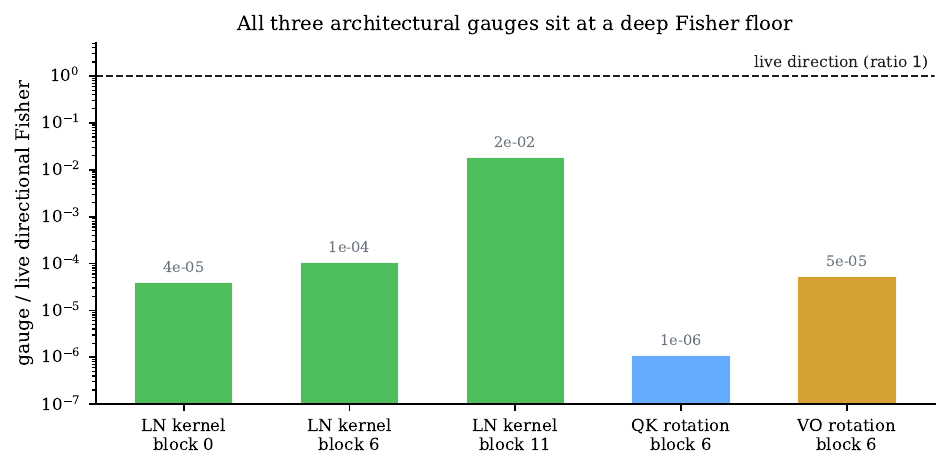}
\caption{The gauge floors on the fine-tuned DINOv2 ViT-S/14. The LayerNorm kernel and the attention query--key rotation read a directional Fisher orders of magnitude below a live direction (ratio $1$); the deep floor is the flat signature the read classifies them by.}
\label{fig:vit-gauge}
\end{figure}

\paragraph{The from-scratch model forms a node-death off the axis.} Trained from scratch
under weight decay, the same architecture prunes its over-parametrised MLP as it
compresses and leaves the dead subspace rotated off the coordinate axes (squared-ReLU
coordinate-concentration $0.07$ at the deepest singular block). A per-coordinate scan
misses such a death; the off-canonical read recovers the activation-predicted order on
all four conditions (Table~\ref{tab:vit}), squared-ReLU $\hat k=3.00$ and gelu
$\hat k\approx2.0$, across two decay strengths and three seeds.

\subsection{Developmental emergence of the dead subspace}
\label{app:developmental}
\begin{figure}[ht]
\centering
\includegraphics[width=0.62\textwidth]{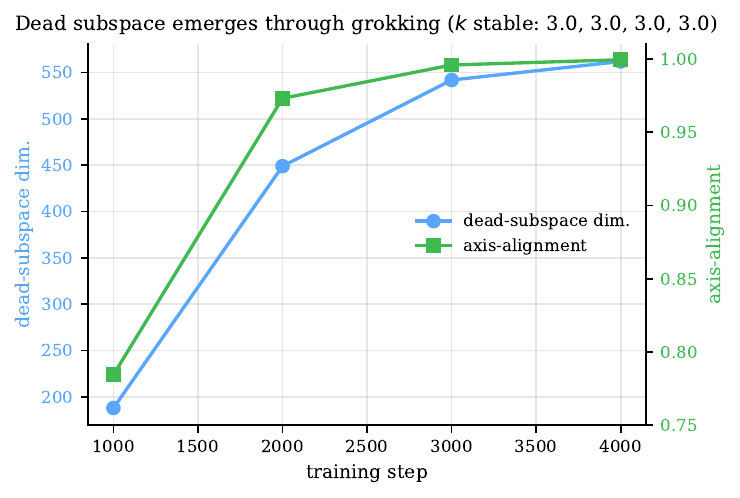}
\caption{The dead subspace emerges through grokking (gauge-fixed d=8). The
dead-subspace dimension (blue) and the axis-alignment (green) both grow across the grok
transition while the recovered order stays at $k\approx3$.}
\label{fig:developmental}
\end{figure}

The frozen-checkpoint read is cheap enough to run at every checkpoint, which traces
the dead subspace through training. Table~\ref{tab:developmental} and
Figure~\ref{fig:developmental} read the gauge-fixed d=8 squared-ReLU run at four
training steps across the grok transition. The dead subspace grows and sharpens onto
the coordinate axes as the network compresses: the dead-subspace dimension rises from $188$ to
$562$ and the axis-alignment from $0.78$ to $0.999$, while the recovered order holds
at $k\approx3$ throughout. The order is fixed by the activation before the subspace
forms; only the count and the alignment develop. This gives the classification a
developmental signature beyond the single-checkpoint read: the architectural gauge
directions are present from initialisation, whereas the node-deaths appear only as
the network compresses.

\begin{table}[ht]\centering\small
\caption{Developmental read of the gauge-fixed d=8 squared-ReLU run (cell
\texttt{gauge\_mdrift0p4\_s42}, block $1$, $n/d\approx64$), at four steps across the
grok transition. The order $k$ is steady while the dead-subspace dimension and axis-alignment
grow.}
\label{tab:developmental}
\begin{tabular}{rrrr}
\toprule
step & dead-subspace dim. & axis-alignment & $\hat k$ \\
\midrule
$1000$ & $188$ & $0.78$  & $2.97$ \\
$2000$ & $449$ & $0.97$  & $2.97$ \\
$3000$ & $542$ & $0.996$ & $2.96$ \\
$4000$ & $562$ & $0.999$ & $3.05$ \\
\bottomrule
\end{tabular}
\end{table}

 \section{Optimizer and basis shape the read}\label{app:sec:shapes}

\subsection{The orthogonaliser base produces the readable basis}
\label{app:basis}
The optimiser claim of Section~\ref{sec:optimizer} reads the deep d=8 squared-ReLU
transformer under three optimisers from one matched cohort, which separates the
orthogonaliser from the gauge projection. The arms are vanilla Muon (textbook NS5, the
\texttt{ns\_off} baseline), the scaled-polar orthogonaliser with the gauge removed (the
\textsc{DDCMuon} orthogonaliser, \texttt{bf\_fast}), and the gauge-equivariant optimiser
on that orthogonaliser (\texttt{bf\_fast} plus the rotation gauge). This is the
dead-structure readability counterpart of the four-arm accuracy decomposition of
\citet{DDCRef}, run on the same cohort. Each block's dead subspace is read by the
multi-component scan: the top eight directions of the dual-factor bottom subspace are
each scanned for the asymptotic $t^{2(k-1)}$ growth, and the count that reaches it (the
\emph{readable} count) separates a formed dead structure from a diffuse one
(Table~\ref{tab:basis}).

Vanilla Muon leaves no readable dead subspace: the bottom block is a small flat
near-kernel (dead-subspace dimension $9$ to $32$), none of whose directions reach the asymptotic
regime, at every interior block and every seed (block-1 readable $0/24$ across three
seeds; blocks $2$ to $4$ at seed $42$ read $0/8$ with axis-alignment falling to $0.10$
to $0.40$). The scaled-polar orthogonaliser, with the gauge removed, instead forms a
large readable subspace (dead-subspace dimension $606$ to $809$) whose directions read the
squared-ReLU order $k\approx3$ on the coordinate axes (axis-alignment $>0.99$), across
three seeds and the interior blocks ($23/24$ readable at block $1$, $8/8$ at each of
blocks $2$ to $4$). The gauge on the same orthogonaliser reads the same order at the
same alignment ($22/24$ at block $1$, $8/8$ through depth), and in addition spectrally
separates the bottom block (block verdict ``subspace'' against the orthogonaliser's
``flat''). So on this deep transformer the readable, axis-aligned dead structure comes from the
scaled-polar orthogonaliser; the gauge's separate contribution here is the spectral
separation of the block. The task-natural reproduction below confirms the orthogonaliser
supplies the alignment at width $128$ as well, once the gauge's weight decay is matched.

The $0/24$ for vanilla Muon is the diffuse-structure verdict: the bottom block forms no
direction that reaches the asymptotic regime, the null the gate returns in
Section~\ref{sec:optimizer}, with nothing formed for any scan to read. This cohort is
squared-ReLU, whose readable structure the orthogonaliser leaves on the coordinate axes, so the
off-canonical recovery of a \emph{rotated} order is exercised on the gelu cohort, where the same
orthogonaliser leaves the order rotated off the axes and the joint-mode scan still returns it
(Figure~\ref{fig:hero}, Section~\ref{sec:optimizer}).

At depth $24$ the diffuse-versus-readable split holds with a thinner signal. That cohort
carries only the vanilla and gauge arms, with no gauge-free scaled-polar cell, so the
orthogonaliser and the gauge cannot be separated there. Its dead subspace is one- to
four-dimensional, so the read is single-component, but vanilla Muon reads no order at any
block under either activation (readable $0/1$ to $0/4$) while the gauge arm reads it on
most blocks (squared-ReLU $3/3$, gelu $2/3$). The contrast is the same as at depth $8$,
and isolating the orthogonaliser from the gauge at width needs a gauge-free scaled-polar
cohort the cohort does not yet contain.

\begin{table}[ht]\centering\small
\caption{Three optimisers from one matched cohort (\texttt{grok\_rope\_ab}, d=8
squared-ReLU, $n/d\approx64$), read by the multi-component scan at block $1$ across
three seeds. The readable count is how many of the top eight dual-factor subspace
directions reach the asymptotic $t^{2(k-1)}$ regime. The readable axis-aligned
structure is the scaled-polar orthogonaliser: both scaled-polar arms form it, vanilla
Muon does not.}
\label{tab:basis}
\setlength{\tabcolsep}{4pt}
\begin{tabular}{l p{3.1cm} l l l l}
\toprule
arm & recipe & block-1 readable & $\hat k$ & dead-subspace dim. & axis-align \\
\midrule
vanilla Muon  & textbook NS5 (\texttt{ns\_off})        & $0/24$  & --            & $9$--$32$    & flat near-kernel \\
scaled-polar  & no gauge (\texttt{bf\_fast})            & $23/24$ & $2.84$--$2.95$ & $606$--$809$ & $0.996$--$0.999$ \\
gauge         & scaled-polar $+$ rotation gauge         & $22/24$ & $2.86$--$2.96$ & $598$--$700$ & $1.000$ \\
\bottomrule
\end{tabular}
\end{table}

\begin{figure}[ht]
\centering
\includegraphics[width=\textwidth]{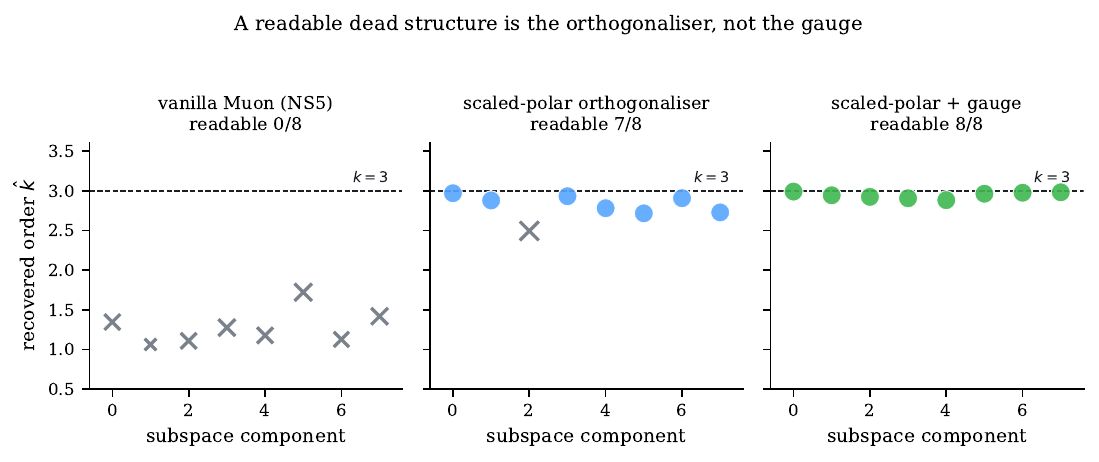}
\caption{The multi-component read at block $1$ (d8 squared-ReLU, seed $42$). Each panel
scans the top eight directions of the dual-factor bottom subspace; a filled marker is an
asymptotic component (the order $\hat k$ it reads), a cross a pre-asymptotic one (no
order). Vanilla Muon forms no readable order; the scaled-polar orthogonaliser and the
gauge both read the squared-ReLU order $k{=}3$ on nearly every component. The readable
dead structure comes from the orthogonaliser; the gauge reads the same.}
\label{fig:readability}
\end{figure}

\paragraph{Task-size corroboration.} The deep-transformer reads move the orthogonaliser and
the depth together. A task-natural one-block transformer (width $128$, the canonical
modular-addition size) repeats the orthogonaliser-versus-gauge contrast and crosses it with
weight decay, five seeds per arm (Table~\ref{tab:basis_tasksize}). The dead subspace is small
here (dead-subspace dimension $1$ to $4$), so this is the axis-alignment read rather than the
multi-component order read the d=8 node-death affords. At matched weight decay the
orthogonaliser lifts the dead-subspace axis-alignment from $0.50$ to $0.69$, and the rotation
gauge adds $0.01$, within seed noise; weight decay leaves the alignment unchanged ($0.50$ at
$1\times$ and $2\times$). The readable axis-aligned basis is therefore the orthogonaliser at
the task size too, matching the deep-transformer read. The grok-speed differences across these
arms are weight-decay effects, accounted for by the DDC optimizer paper \citep{DDCRef}; the
dead-structure read here is the alignment.

\begin{table}[ht]\centering\small
\caption{Task-natural one-block transformer (width $128$, squared-ReLU, five seeds):
dead-subspace axis-alignment crossing the orthogonaliser and the rotation gauge with weight
decay. At matched ($1\times$) weight decay the orthogonaliser supplies the alignment ($+0.19$)
and the gauge adds little ($+0.01$, within noise); weight decay does not move alignment
(vanilla $1\times = 2\times$), the control that isolates the alignment read.}
\label{tab:basis_tasksize}
\begin{tabular}{lll}
\toprule
arm & weight decay & axis-align \\
\midrule
NS5 vanilla                 & $1\times$ & $0.499$ \\
NS5 vanilla                 & $2\times$ & $0.497$ \\
scaled-polar (no gauge)     & $1\times$ & $0.692$ \\
scaled-polar $+$ gauge      & $1\times$ & $0.704$ \\
\bottomrule
\end{tabular}
\end{table}

\begin{figure}[ht]
\centering
\includegraphics[width=0.6\textwidth]{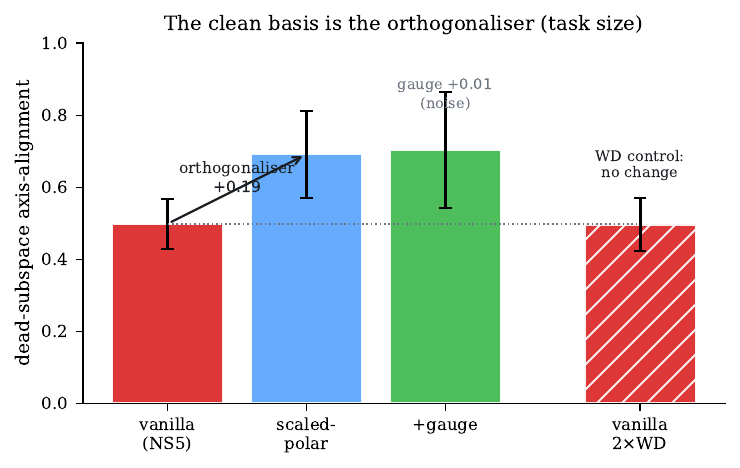}
\caption{The dead-subspace axis-alignment at task size (1-block d128, 5 seeds). At matched
$1\times$ weight decay the alignment jumps from vanilla to the scaled-polar orthogonaliser
($+0.19$) and the rotation gauge adds nothing ($+0.01$, within seed noise), so the clean
axis-aligned basis is the orthogonaliser. The hatched $2\times$-WD vanilla bar is the control:
weight decay does not move the alignment, which isolates the read.}
\label{fig:tasksize}
\end{figure}

\subsection{The optimiser-axis cell}
\label{app:optaxis}
The optimiser-axis reads of Section~\ref{sec:optimizer} (Tables~\ref{tab:optaxis}
and~\ref{tab:noisytarget}) hold a small architecture fixed and move only the
optimiser, isolating the optimiser's effect from the depth the deep-transformer reads
confound it with. The cell is a two-layer squared-ReLU teacher--student MLP (input
dimension $16$, hidden width $64$, output dimension $4$); the teacher uses $4$ hidden
units and the student $64$, so the student over-covers and its spare units prune to a
node-death under weight decay. Each arm trains to convergence under SGD, AdamW,
RMSProp, or Adam at three seeds $\{42,142,242\}$. The read is the same off-canonical
pipeline as the transformer cells: the dead unit is nominated from the smaller-norm
K-FAC factor, and the joint mode is scanned on the frozen-weight true-MC Fisher over
the auto-selected purity window. The gauge companion is a single-sided move at the
dead unit; it reads flat only when the unit's consumer has also been pruned, the
signature of a confirmed node-death.

\begin{figure}[ht]
\centering
\includegraphics[width=\textwidth]{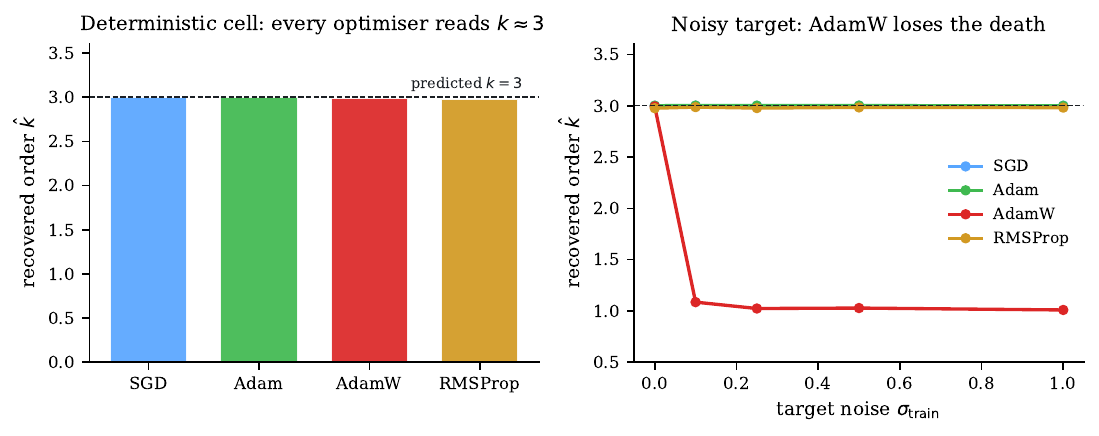}
\caption{The optimiser-axis cell. Left: on the deterministic squared-ReLU teacher--student node-death every optimiser reads the predicted $k{=}3$. Right: under Gaussian target noise AdamW loses the death (the order reads deviant) while SGD, Adam, and RMSProp hold it.}
\label{fig:optaxis}
\end{figure}

Section~\ref{sec:optimizer} reads the deterministic-target and noisy-target results
(Tables~\ref{tab:optaxis} and~\ref{tab:noisytarget}): the order holds at $k\approx3$ where a
clean death forms, the optimiser sets the basis, and a noisy target separates the optimisers
through the adaptive preconditioner's noise-fitting amplification. Figure~\ref{fig:optaxis}
shows the two cells side by side.

\subsection{Dead-input classification on the seven-optimizer cohort}
\label{app:component_death}
A dead direction the read cannot order is the limiting case of the read. The
component is degenerate enough that no perturbation of its weights moves the output,
so the transversal-Fisher scan finds nothing to fit and the order is undefined. This
appendix documents that case, the read's third classification beside the gauge and the
node-death. The cohort is a one-block transformer (width
$128$, four heads) on addition mod $113$, the modular-addition cohort \citet{DDCRef}
analyses for its mechanism, read at the grokked checkpoint ($0.98$ to
$1.00$ validation), across seven optimizers and three activations with three seeds
each. The read targets the smallest-joint-norm hidden unit of the MLP.

On the arm trained by the gauge-equivariant Adam the read returns no order: every
scanned direction classifies tangential and the transversal count is zero. The cause
is upstream. The second LayerNorm gain has collapsed to $3\times10^{-9}$, eight orders
below the $0.20$ to $0.36$ the other arms hold, so the normalised MLP input is a
constant and the unit's pre-activation no longer depends on the token. The liveness
read of the MLP input (one forward pass over
the gain norm and the input-activation spread) reports the collapse, so the bare
no-order result classifies a dead input channel, distinct from a gauge at a deep floor
and a node-death at a finite order. This collapse is the DDC optimizer shedding the
spare feed-forward block on its log-radial scale, where weight decay drives the gauged
scale to zero \citep{DDCRef}; the read's part is to localise it from the frozen
checkpoint, where the same rotation gauge on a Muon base (the DDC arm of
Appendix~\ref{app:ranking}) instead keeps the MLP alive.

The restricted learning coefficient confirms the classification.
The MLP block's SGLD-LLC on the gauge-equivariant Adam arm is near zero, a seed median
of $0.4$ against $17$ to $756$ for the live arms, the cohort floor by more than
an order of magnitude. The same restriction taken to a single dead unit does not
converge (Appendix~\ref{app:llc}); the block is the finest scale the sampler resolves
on this arm, while the rate read returns the per-direction order.

Across the seven optimizers and three seeds the read separates this dead-input arm from
the live-MLP arms at the frozen checkpoint, the classification the cross-optimizer
analyses of Appendices~\ref{app:ranking} and~\ref{app:ranking_perblock} build on.

 \section{The global view}\label{app:sec:global}

\subsection{Learning-coefficient comparison}
\label{app:llc}
Table~\ref{tab:llc} reports the SGLD learning-coefficient estimate at three
granularities against the rate read, on the grokked d=8 squared-ReLU checkpoint. The estimator
is the SGLD local learning coefficient of \citet{LauFurmanWangMurfetWei25}, its restricted
(per-subset) form following \citet{WangHoogland24}, run via the devinterp library
\citep{devinterp} with our own calibration of the inverse temperature and convergence gates.
The global and per-block (restricted) estimates converge at component scale; the
per-direction restriction does not converge. The rate read returns the
per-direction order deterministically.

\begin{table}[ht]\centering\small
\caption{Learning-coefficient estimate vs the rate read (grokked d=8 checkpoint).}
\label{tab:llc}
\setlength{\tabcolsep}{4pt}
\begin{tabular}{l l l c p{3.5cm}}
\toprule
measure & granularity & value & $\hat R$ & cost \\
\midrule
SGLD-LLC (calibrated)        & global model   & $\lambda{=}116.7\pm2.2$ & $0.99$ & $5{\times}5$ grid $+$ estimate ($453$\,s) \\
SGLD-LLC (locked, gelu cell) & global model   & $\lambda{=}157\pm57$    & $1.30$ & locked config does not transfer \\
restricted SGLD-LLC          & one MLP block   & $\lambda{=}147\pm2$     & $0.99$ & one estimate \\
restricted SGLD-LLC          & one dead unit   & degenerate ($<0$)       & --     & no calibration setting passes \\
rate read (this work)        & one direction   & $k{=}3$, $\lambda_{\mathrm{dir}}{=}1/6$ & -- & one $18$\,s pass, deterministic \\
\bottomrule
\end{tabular}
\end{table}

\begin{figure}[ht]
\centering
\includegraphics[width=0.7\textwidth]{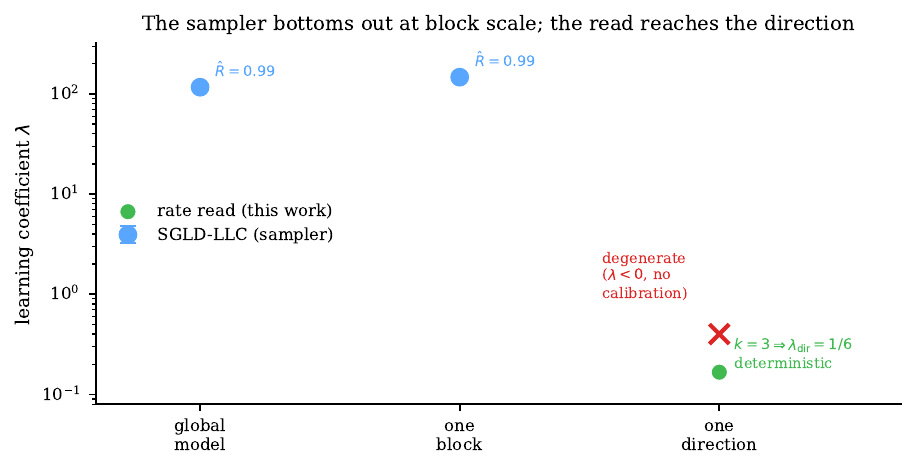}
\caption{The learning coefficient across granularities (grokked d8 squared-ReLU). The SGLD sampler converges at the global model and one block but degenerates restricted to a single direction; the descent-free rate read returns the per-direction order $k{=}3$, hence $\lambda_\mathrm{dir}{=}1/6$, deterministically.}
\label{fig:llc}
\end{figure}

On the seven-optimizer width-128 cohort of Appendix~\ref{app:component_death} the read
returns the three views of the triple at one frozen checkpoint
(Table~\ref{tab:triple}): the per-direction order $k$ and its coefficient $1/(2k)$ from the
geometry read, the block coefficient from the restricted sampler
(Appendix~\ref{app:ranking_perblock}), and the trajectory fluctuation $\nu$ from the loss
curve. The views are complementary. The order separates the geometries, the canonical arms
reading the architecture order $3$, the weight-decay arms departing to the input-death order
near $2$, and the gauge-equivariant Adam arm returning no order with a dead input
(Appendix~\ref{app:component_death}). The block coefficient carries the bulk degeneracy a
single direction does not: SGD and AdamW share the order $3$ yet differ thirtyfold in the
restricted coefficient, the dense weight-decay-free posterior against the compressed one.
The fluctuation is read on the trajectory the geometry read does not need.

\begin{table}[ht]\centering\small
\caption{The Watanabe triple read three ways on the seven-optimizer cohort, seed median,
squared-ReLU. Order $k$ is the intrinsic geometry read (the weight-decay arms read the
off-canonical input-death order, gauge Adam's input is dead with no order, the DDC
log-radial shed of Appendix~\ref{app:component_death}); $1/(2k)$ the
per-direction coefficient; MLP rLLC the local learning coefficient from the sampler restricted to the block
(Table~\ref{tab:perblock}); $\nu$ the trajectory fluctuation at the plateau.}
\label{tab:triple}
\begin{tabular}{lrrrr}
\toprule
optimizer & order $k$ & $1/(2k)$ & MLP rLLC & $\nu$ \\
\midrule
SGD            & $3.0$  & $0.17$ & $756$ & $20$  \\
Muon           & $1.9$  & $0.26$ & $17$  & $6$   \\
Muon$+$WD      & $2.0$  & $0.25$ & $27$  & --    \\
AdamW          & $3.0$  & $0.17$ & $23$  & $28$  \\
AdamW matched  & $3.0$  & $0.17$ & $55$  & $10$  \\
gauge Muon     & $2.0$  & $0.25$ & $66$  & $7$   \\
gauge Adam     & dead   & --     & $0.4$ & $108$ \\
\bottomrule
\end{tabular}
\end{table}

\subsection{Complexity ranking against the sampler}
\label{app:ranking}
This appendix backs the ranking use of Section~\ref{sec:uses}
(Figure~\ref{fig:complexity_ranking}). The set is eighteen squared-ReLU d8
checkpoints on one synthetic arithmetic task (the task and architecture fixed),
spanning gauge-fixed DDC, vanilla Muon, and vanilla AdamW, with and without weight
decay, across training steps. For each the cheap estimate is the dead-unit census
(one forward pass), summed into
$\hat\lambda = N_{\mathrm{total}}/2 - N_{\mathrm{dead}}(1/2 - 1/(2k))$ over the
$N_{\mathrm{total}}=16384$ MLP-hidden units with $k=3$; the expensive estimate is
the calibrated SGLD-LLC with a focused
$(\eta,\gamma)$ grid per cell, since one shared configuration does not converge
positively across the set. Eleven cells converge ($\hat R \le 1.1$, positive
estimate); the other seven drop from the SGLD comparison on a negative estimate or
$\hat R>1.1$ (the four no-weight-decay accumulation-regime vanilla-Muon cells, a late
vanilla-Muon cell, one gauge-fixed DDC cell, and the vanilla-AdamW cell), while the
census reads all eighteen. Spearman $\rho(\hat\lambda, \text{SGLD-LLC}) = 0.66$ on the
eleven converged checkpoints and $0.71$ on the seven distinct converged models
(Table~\ref{tab:ranking}). The gauge-fixed DDC models occupy the high-$\hat\lambda$,
high-LLC corner and the vanilla-Muon models the low corner, in both estimators. The
census separates the families by the dead-unit count, which is $0.1\%$ to $4.4\%$ of
the $16384$ units, so $\hat\lambda$ barely moves off $N_{\mathrm{total}}/2$: the pooled
$\rho$ reads the family split, while within a single run the count follows the
weight-decay schedule and the sampler moves the other way. The per-block effective
dimension below tracks the restricted sampler more closely.

\begin{table}[ht]\centering\small
\caption{The seven distinct converged models, sorted by the cheap $\hat\lambda$.
Both estimators place the gauge-fixed DDC models above the vanilla-Muon models.
Seed $42$; $n{=}960$ validation sequences; per-cell SGLD calibration over
$\eta\in\{1,3,10\}{\times}10^{-5}$, $\gamma\in\{10,100\}$.}
\label{tab:ranking}
\begin{tabular}{llrrl}
\toprule
model & optimizer & $N_{\mathrm{dead}}$ & cheap $\hat\lambda$ & SGLD-LLC ($\hat R$) \\
\midrule
DDC 1     & DDC          & $17$  & $8186$ & $121.1{\pm}4.0$ ($1.005$) \\
DDC 2     & DDC          & $28$  & $8183$ & $151.4{\pm}3.6$ ($0.994$) \\
DDC 3     & DDC          & $29$  & $8182$ & $99.9{\pm}2.6$ ($1.009$) \\
DDC 4     & DDC          & $99$  & $8159$ & $99.7{\pm}8.9$ ($1.049$) \\
DDC 5     & DDC          & $106$ & $8157$ & $133.1{\pm}2.8$ ($1.003$) \\
Muon      & vanilla Muon & $508$ & $8023$ & $77.3{\pm}24.2$ ($1.022$) \\
Muon$+$WD & vanilla Muon & $721$ & $7952$ & $41.8{\pm}14.2$ ($1.008$) \\
\bottomrule
\end{tabular}
\end{table}

\subsubsection{Per-block ranking on a seven-optimizer cohort}
\label{app:ranking_perblock}
The ranking above reads the cheap census against the global coefficient. The same question
at the block scale, on the width-128 seven-optimizer cohort of
Appendix~\ref{app:component_death} (SGD, Muon, Muon with weight decay, AdamW, AdamW with
matched hyperparameters, gauge-fixed Muon, gauge-fixed Adam; squared-ReLU, three seeds),
reads the calibrated SGLD coefficient restricted to the MLP block and to the attention
block (over the block parameter subset, seed median
under a cross-seed stability gate) against a cheap
frozen-Fisher proxy. The proxy is the localized effective dimension
$\hat\lambda_{\mathrm{eff}}(\gamma) = \tfrac12 \sum_i \lambda_i/(\lambda_i + \gamma)$ over
the block's K-FAC factor spectrum (the products $\lambda_{G,i}\lambda_{A,j}$ of the input
and output-gradient factor eigenvalues), at one
forward and backward pass per block. Table~\ref{tab:perblock} lists the seed medians.

\begin{table}[ht]\centering\small
\caption{Per-block complexity on the seven-optimizer cohort: the calibrated SGLD coefficient
restricted to each block (rLLC) against the cheap K-FAC effective dimension at $\gamma=0.1$.
Seed medians, squared-ReLU. The gauge-Adam MLP is the DDC log-radial shed
(Appendix~\ref{app:component_death}); the read and the restricted sampler both localise it.}
\label{tab:perblock}
\begin{tabular}{lrrrr}
\toprule
optimizer & MLP rLLC & MLP $\hat\lambda_{\mathrm{eff}}$ & ATTN rLLC & ATTN $\hat\lambda_{\mathrm{eff}}$ \\
\midrule
SGD            & $756$ & $65$  & $624$ & $191$ \\
Muon           & $17$  & $9$   & $175$ & $22$  \\
Muon$+$WD      & $27$  & $6$   & $146$ & $14$  \\
AdamW          & $23$  & $20$  & $352$ & $140$ \\
AdamW matched  & $55$  & $12$  & $634$ & $198$ \\
gauge Muon     & $66$  & $49$  & $207$ & $122$ \\
gauge Adam     & $0.4$ & $0.0$ & $230$ & $258$ \\
\bottomrule
\end{tabular}
\end{table}

The effective dimension tracks the restricted sampler per block: Spearman
$\rho(\hat\lambda_{\mathrm{eff}}, \mathrm{rLLC}) = 0.82$ on the MLP and $0.79$ on the
attention, at $\gamma = 0.1$, near the localization scale the sampler itself uses. The bare
factor rank does not track ($\rho = 0.29$): SGD carries a low-rank MLP input from the
low-dimensional grokked representation yet the cohort's highest MLP coefficient from its
dense weight-decay-free posterior, so the directions have to be weighted by curvature, which
the effective dimension does and a count does not. The liveness-gated census
reads the input-dead arm to zero
(gauge Adam, $\hat\lambda_{\mathrm{gated}} = 0$ against $94$ to $256$ for the live arms), the
same death the restricted sampler reads ($0.4$). The attention block carries no isolated
dead channels; its degeneracy is a low weight-space rank, and the per-head VO and QK
effective rank separates the optimizer families the restricted sampler does, the gauge and
vanilla Muon arms at high weight-rank and low coefficient, the Adam arms at low rank and high
coefficient.

\subsection{Global-coefficient assembly on analytic models}
\label{app:global}
Table~\ref{tab:global} assembles a global learning coefficient from the
per-direction reads on analytic singular models and checks it against the closed
form. For each cell the per-direction orders come from the
multi-component read, the crossing-versus-independent grouping
from the directional-Fisher coupling (two
directions share a locus when fixing one scales the other's leading Fisher
coefficient, an exponent of $2k_j$ within a crossing and $0$ across independent
loci), and the global pair from the sum-versus-crossing rule. The closed form is Watanabe's normal-crossing
real log canonical threshold (RLCT) for the crossings, the separable sum for the independent loci, and Aoyagi's
deep-linear coefficient for the scalar networks. On the scalar networks the radial
collapse direction reads order $k=L$, whose single-direction threshold $1/(2L)$ sits
below the global $1/2$ the $L$-hyperplane crossing carries; the blind grouping
recovers the $1/2$.

The wide deep linear networks mark the boundary of the read. Their singular set is
the matrix product-zero locus, a determinantal variety whose crossings lie in
coupled directions away from any coordinate. The per-direction coordinate read
departs from the closed form: on the small grids it over-reads to the regular $D/2$
at depth two, each coordinate staying regular at order one, and under-reads to $1/2$
at depth three, the grouping collapsing every coordinate into a single crossing
(lower block of Table~\ref{tab:global}). The improved rotated order read leaves this
unchanged, so the gap is in the assembly. Three reads survive the boundary. The
Newton polyhedron of the KL divergence's monomial support gives the learning
coefficient by a toric computation, the diagonal Newton distance solved as a linear
program: exact for a singularity non-degenerate with respect to its Newton
polyhedron, reproducing the closed form on every enumerable cell, and the simplex
upper bound otherwise. On the determinantal cells the simplex bound sits far below
the regular $D/2$ (the depth-two $2{\times}2{\times}2$ reads $2.0$ against $D/2 = 4.0$
and the exact $1.5$) and is itself exact on some cells ($3{\times}2{\times}2{\times}3$
reads the exact $2.0$). The generic-line order $k_{\mathrm{gen}}$ then brackets the
coefficient as $[1/(2 k_{\mathrm{gen}}),\, \min(D/2,\, \text{simplex})]$, which
contains the closed-form $\lambda$ on every cell of the table and holds the
hundreds-scale $\lambda$ at the trained $H=(20,h,h,20)$ for $h$ up to $128$. The
determinantal signature is itself detectable: the generic-line order disagrees with
the order the recovered coordinate structure implies, which flags the locus blind and
recovers its depth $L$. Routing the flagged locus to the structured resolution rule,
Aoyagi's coefficient for the architecture, recovers the closed form on all five cells
the blind assembly missed, the determinantal resolution \citet{TheoryRefNamed} proves for
depth $L\le3$. A general exact resolver for an arbitrary determinantal
variety stays open in that programme: the simple origin blow-up \citep{Hironaka64} stalls at the simplex bound,
so the exact value needs the structured rank-locus resolution (Section~\ref{sec:discussion}).

\paragraph{Typing the intersection.} The coupling grouping generalises to a per-cell read of
the intersection type, which routes the assembly to its matched rule: a transversal crossing of
regular loci to the minimum, a separable sum to the sum, a
determinantal locus to the structured resolution, and a tangency to the Newton-polyhedron
engine. A tangency and a transversal crossing can read the same generic order; the offset
resolution separates them, since holding one nominated normal at an offset makes a transversal
crossing's per-direction orders sum to the generic order while a tangency's overshoot it, no
offset peeling a clean factor off a shared tangent. On the constructed tangency
$K=(y^2-x^4)^2$ this routes to Newton for the exact $3/8$ the crossing rule misses, and the
matched rule reproduces the closed form on every typed cell. The tangency whose loci read order
one along the shared tangent, where the offset orders do not overshoot and a line scan reads it
as a transversal crossing, separates under a second-order read. Near the singular point the
directional Fisher grows as $t^2(u^\top A v)^2$ with $A$ the Hessian of the locus, and the rank
of $A$ types the cell: a transversal crossing reads full rank, every sheet entering at second
order, and a tangency reads rank-deficient, the contact pushing one direction to a higher order.
The confidence is the singular-value gap, which falls toward the decision threshold as a
crossing approaches a shallow angle and the tangency boundary. The case still open is a contact
tangent to second order, where $A$ itself degenerates and the read needs a third order.

\begin{table}[ht]\centering\small
\caption{Assembled global coefficient against the closed form on analytic models,
with the rigorous bracket and the resolution-rule value where the assembly fails.
Orders are recovered blind; the locus grouping is recovered from the Fisher coupling;
the assembly is the sum-versus-crossing rule (reproduced exactly on the enumerable
block by the Newton-polyhedron engine); the bracket is
$[1/(2 k_{\mathrm{gen}}),\, \min(D/2, \text{simplex})]$, the determinantal upper
bound being the Newton simplex bound. The resolved column is the structured
resolution rule (Aoyagi) the determinantal detector routes to. Seed $42$,
$n{=}2{\times}10^{4}$ Monte-Carlo for the directional Fisher (the wide block is exact
Gauss-Newton). Assembled equals the closed form on every enumerable cell (upper
block); on the wide matrix determinantal cells (lower block) the assembly departs,
the bracket contains the closed form, and the resolution rule recovers it.}
\label{tab:global}
\setlength{\tabcolsep}{4pt}\begin{tabular}{llllll}
\toprule
cell & rec.\ $k$ & assembled $(\lambda, m)$ & bracket & resolved $(\lambda, m)$ & closed form $(\lambda, m)$ \\
\midrule
crossing $(2,2)$   & $2,2$        & $(1/4,\,2)$   & $[0.12,\,1.00]$ & --           & $(1/4,\,2)$ \\
crossing $(2,3)$   & $2,3$        & $(1/6,\,1)$   & $[0.10,\,1.00]$ & --           & $(1/6,\,1)$ \\
crossing $(2,3,2)$ & $2,3,2$      & $(1/6,\,1)$   & $[0.07,\,1.50]$ & --           & $(1/6,\,1)$ \\
crossing $(3,3,3)$ & $3,3,3$      & $(1/6,\,3)$   & $[0.06,\,1.50]$ & --           & $(1/6,\,3)$ \\
sum $(2,3)$        & $2,3$        & $(5/12,\,1)$  & $[0.25,\,1.00]$ & --           & $(5/12,\,1)$ \\
sum $(2,3,4)$      & $2,3,4$      & $(13/24,\,1)$ & $[0.25,\,1.50]$ & --           & $(13/24,\,1)$ \\
composite          & $2,2,3$      & $(5/12,\,2)$  & $[0.17,\,1.50]$ & --           & $(5/12,\,2)$ \\
composite          & $2,3,2,2$    & $(5/12,\,2)$  & $[0.12,\,2.00]$ & --           & $(5/12,\,2)$ \\
DLN $L{=}2$        & $1,1$        & $(1/2,\,2)$   & $[0.25,\,1.00]$ & --           & $(1/2,\,2)$ \\
DLN $L{=}3$        & $1,1,1$      & $(1/2,\,3)$   & $[0.17,\,1.50]$ & --           & $(1/2,\,3)$ \\
DLN $L{=}4$        & $1^{\times 4}$ & $(1/2,\,4)$ & $[0.12,\,2.00]$ & --           & $(1/2,\,4)$ \\
DLN $L{=}5$        & $1^{\times 5}$ & $(1/2,\,5)$ & $[0.10,\,2.50]$ & --           & $(1/2,\,5)$ \\
\midrule
\multicolumn{6}{l}{\textit{Wide matrix deep linear (matrix product-zero, determinantal): the boundary}} \\
DLN $2{\times}2{\times}2$         & $1^{\times 8}$  & $(4,\,1)$    & $[0.25,\,2.00]$ & $(3/2,\,1)$ & $(3/2,\,1)$ \\
DLN $3{\times}2{\times}3$         & $1^{\times 12}$ & $(6,\,1)$    & $[0.25,\,3.00]$ & $(5/2,\,1)$ & $(5/2,\,1)$ \\
DLN $4{\times}2{\times}4$         & $1^{\times 16}$ & $(8,\,1)$    & $[0.25,\,4.00]$ & $(7/2,\,1)$ & $(7/2,\,1)$ \\
DLN $3{\times}3{\times}3$         & $1^{\times 18}$ & $(9,\,1)$    & $[0.25,\,4.50]$ & $(7/2,\,2)$ & $(7/2,\,2)$ \\
DLN $3{\times}2{\times}2{\times}3$ & $1^{\times 16}$ & $(1/2,\,16)$ & $[0.17,\,2.00]$ & $(2,\,3)$  & $(2,\,3)$ \\
\bottomrule
\end{tabular}
\end{table}

\begin{figure}[ht]
\centering
\includegraphics[width=\textwidth]{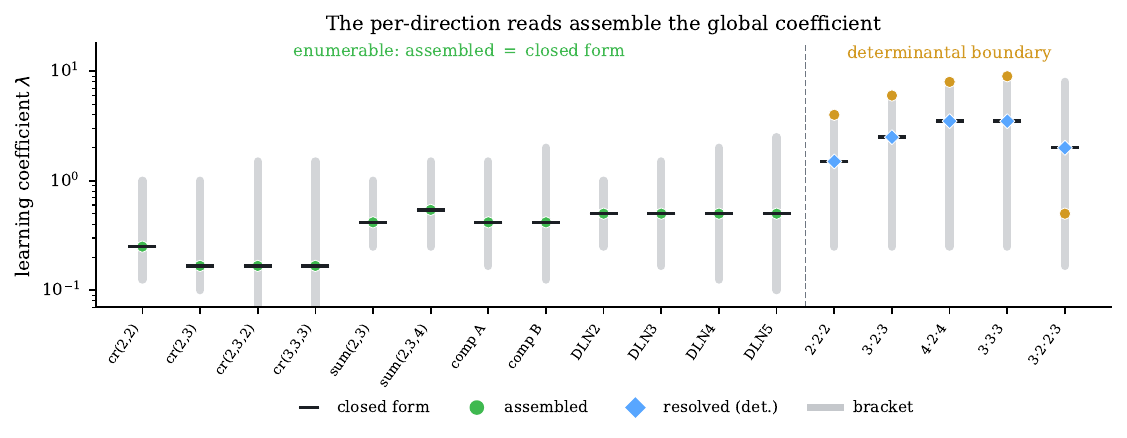}
\caption{Assembling the global learning coefficient against the closed form. The per-direction reads assemble to the closed form (black tick) on every enumerable cell (green) and depart on the wide matrix determinantal cells (amber, right of the divider); the rigorous bracket (grey) contains the closed form and the determinantal-routed resolution (blue) recovers it.}
\label{fig:global}
\end{figure}

\subsection{The singular-fluctuation cells}
\label{app:nu}
Section~\ref{sec:trajectory} reads the singular fluctuation $\nu$ from the order through the
universality value $\nu(k)$ and measures where a trained network's realized $\nu$ falls below it.
Three cells support that section, all CPU and fp64, seed $42$.

The \emph{order-$k$ validation cell} confirms $\hat\nu=\nu(k)$ on an isolated direction. The
model is $y=a\,s^{k}+\varepsilon$, $\varepsilon\sim\mathcal{N}(0,\sigma^{2})$ ($a{=}1$,
$\sigma{=}1$), whose KL along $s$ has order $2k$. We form the exact one-dimensional posterior on
a grid and compute the functional variance $V_n=\sum_i\mathrm{Var}_{\mathrm{post}}[\log
p(y_i\mid s)]$, with $\hat\nu=V_n/2$, data-averaged over $300$ draws at each
$n\in\{500,1000,2000,4000\}$.

The \emph{estimator anchor} validates the functional-variance estimator on a regular
$d$-parameter linear-Gaussian model, where $\nu=d/2$. The closed form and a hand-rolled SGLD
sampler of the same posterior both return $d/2$ ($d=4,6,10\to2.0,3.0,5.0$).

The \emph{absorption cell} isolates the live-structure suppression. The model is
$y=b\,g(x)+s^{k}c(x)+\varepsilon$ with a regular parameter $b$ (basis $g$) and an order-$k$ dead
coordinate $s$ (basis $c$), the teacher setting $s^{*}{=}0$. The dead direction's contribution to
$\nu$ is $\nu(\text{joint over }b,s)-\nu(b\text{ alone})$, sampled from the $(b,s)$ grid posterior
and data-averaged over $120$ draws, swept over the basis overlap $\rho=\mathrm{corr}(g,c)$
(Figure~\ref{fig:nu-absorption}). The effective overlap $\rho_{\mathrm{eff}}$ of a generic dead-scan
basis with the live-unit span in the trained vision-transformer cells is the mean over $200$
random scan directions.
 
\end{document}